\pdfoutput=1
\documentclass[11pt]{article}
\usepackage[preprint]{acl}
\usepackage[T1]{fontenc}
\usepackage[utf8]{inputenc}
\usepackage{times,latexsym,microtype,inconsolata}
\usepackage{graphicx} \graphicspath{{figures/}}
\usepackage{amsmath,amssymb,mathabx,mathtools,amsthm,nicefrac}
\usepackage[capitalise,nameinlink]{cleveref}
\usepackage[skip=3pt,font=small]{subcaption}
\usepackage[skip=3pt,font=small]{caption}
\usepackage[printonlyused]{acronym}
\usepackage[linesnumbered,ruled,vlined]{algorithm2e}
\usepackage{wrapfig}
\usepackage[dvipsnames,svgnames,x11names,table]{xcolor}
\usepackage{enumerate}
\usepackage{enumitem}
\usepackage{xspace}
\usepackage[misc]{ifsym}
\usepackage{titletoc}
\usepackage{verbatim,fancyvrb,fvextra,listings}
\usepackage{booktabs,tabularx,colortbl,multirow,multicol,array,makecell,tabularray}
\usepackage{siunitx}

\makeatletter
\DeclareRobustCommand\onedot{\futurelet\@let@token\@onedot}
\def\@onedot{\ifx\@let@token.\else.\null\fi\xspace}
\def\eg{\emph{e.g}\onedot} 

\def\ie{\emph{i.e}\onedot}

\def\vs{\emph{vs}\onedot}

\makeatother

\acrodef{sota}[SOTA]{State-of-the-Art}
\acrodef{method}[\textsc{PRA}]{Preference-based Robot Assistant}
\acrodef{pbp}[\textsc{PbP}]{Preference-based Planning}
\acrodef{vln}[VLN]{Vision-and-Language Navigation}
\acrodef{llm}[LLM]{Large Language Model}
\acrodef{EILEV}[EILEV]{Efficient In-context Learning on Egocentric Videos}
\acrodef{vlm}[VLM]{Vision-Language Model}
\acrodef{vivit}[ViViT]{Video Vision Transformer}
\acrodef{llava}[LLaVA]{Large Language and Vision Assistant}
\acrodef{ai}[AI]{Artificial Intelligence}
\acrodef{ik}[IK]{Inverse Kinematics}
\acrodef{tom}[ToM]{Theory of Mind}
\acrodef{ompl}[OMPL]{Open Motion Planning Library}
\acrodef{sem}[SEM]{Structural Equation Model}
\acrodef{rlhf}[RLHF]{Reinforcement Learning from Human Feedback}
\acrodef{intu}[InTU]{Inferring the Unspoken}

\newcolumntype{x}{>{\columncolor{LightCyan1}}c}
\newcolumntype{y}{>{\columncolor{MistyRose}}c}
\title{Inferring the Unspoken:\\Aligning Embodied Agents with Implicit Preferences}

\author{%
    \textbf{Manjie Xu} \textsuperscript{1,2,3,5,7,\textsuperscript{*}} 
    \textbf{Xinyi Yang} \textsuperscript{1,2,3,5,7,\textsuperscript{*}}
    \textbf{Wei Liang} \textsuperscript{4,$^{\,\textrm{\Letter}}$} 
    \textbf{Chi Zhang} \textsuperscript{3,5$^{\,\textrm{\Letter}}$}
    \textbf{Yixin Zhu} \textsuperscript{2,1,5,6,7,8$^{\,\textrm{\Letter}}$}\\
    \small *\,Equal contribution \quad \textrm{\Letter} Corresponding author \\
    \small\textsuperscript{1} Institute for Artificial Intelligence, Peking University \\
    \small\textsuperscript{2} School of Psychological and Cognitive Sciences, Peking University \\
    \small\textsuperscript{3} School of Intelligence Science and Technology, Peking University
    \small\textsuperscript{4} Beijing Institute of Technology\\
    \small\textsuperscript{5} State Key Lab of General AI, Peking University \quad 
    \small\textsuperscript{6} PKU-Wuhan Institute for Artificial Intelligence\\
    \small\textsuperscript{7} Beijing Key Laboratory of Brain-Computer Interface and Mental Health Modulation \quad 
    \small\textsuperscript{8} XSVision\\
    \small\url{https://sites.google.com/view/preference-based-planning} \\
}

\begin{document}
\maketitle

\vspace{6pt}

\begin{abstract}
Instructions are never complete specifications of the behavior they request: speakers say only what their listeners cannot supply for themselves \citep{grice1975logic,zipf2016human}.
An agent told to ``prepare an apple'' must decide unaided whether to wash or cut it and where to place it. Such choices differ between people and are seldom stated, only enacted \citep{slovic1995construction,lichtenstein2006construction}. Multimodal foundation models render instructions into grounded plans \citep{brohan2023can,driess2023palm}, yet settle these residual choices by generic priors rather than by evidence about the person served \citep{zhang2024building}.
Whether an agent can instead recover such constraints from prior behavior has not been separated from the easier question of planning once they are supplied.
Here we show the failure to be largely one of acquisition rather than planning, and that stating the inferred preference before acting recovers much of what is lost.
\acf{pbp} benchmarks 290 preferences over action parameters, placement policies, and temporal orderings, asking agents to infer from a few unlabeled demonstrations which constraint governs a new underspecified instruction. Given the preference, the strongest models plan categorical tasks at near-zero error; required to infer it, performance drops sharply, and without demonstrations accuracy falls by over a third.
Because a verbalized rule keeps intent while discarding perceptual particulars, preferences so represented transfer to unfamiliar scenes, whereas implicit imitation stays bound to the rooms it was learned in.
The bottleneck lies not in generating actions but in the pragmatic inference preceding them, for which language proves an effective medium.
\end{abstract}

\acresetall

\section{Introduction}

\begin{figure*}[t!]
    \centering
    \small
    \includegraphics[width=\linewidth]{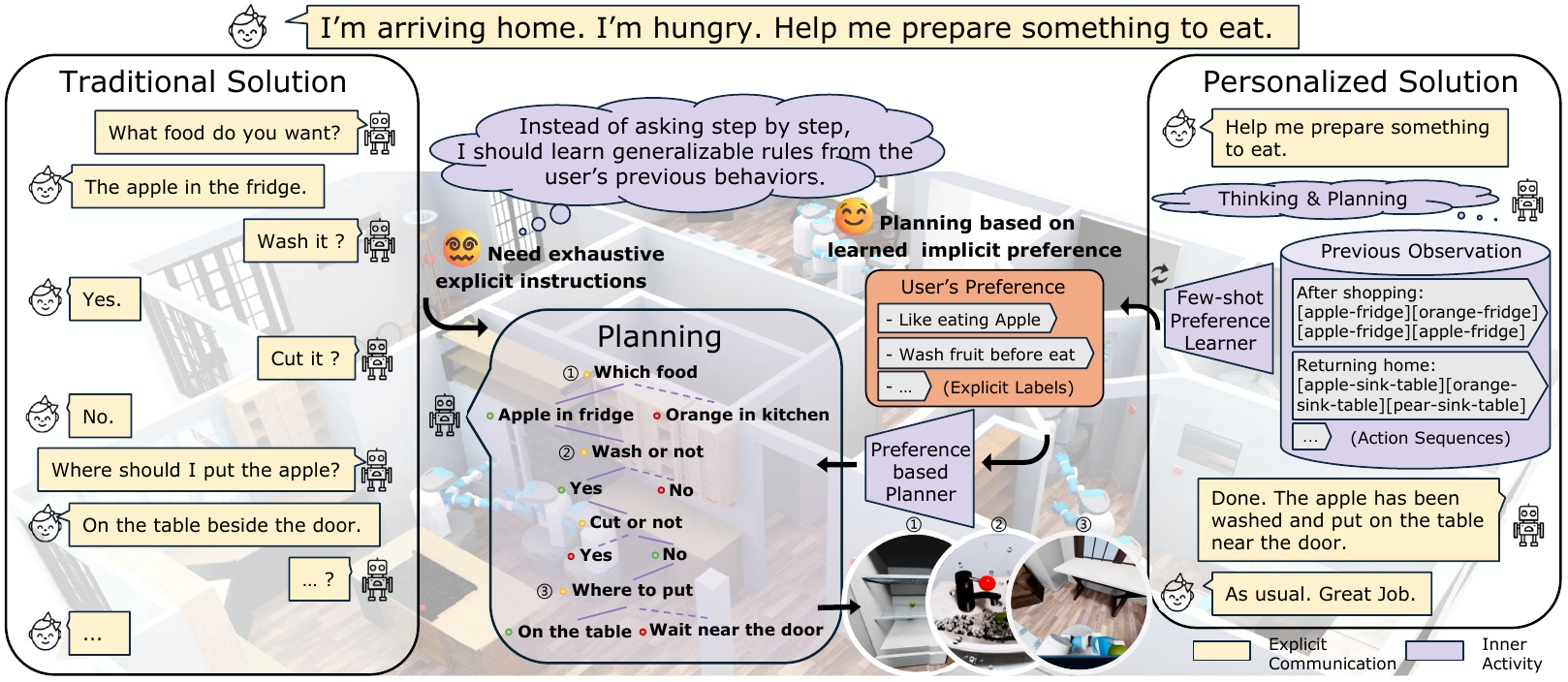}
    \caption{\textbf{The pragmatic gap in embodied planning.} An underspecified instruction such as ``Prepare food'' settles neither which item to use, nor whether to wash or cut it, nor where to put it down (center). A conventional agent recovers these one at a time by interrogating the user (left), whereas \acs{intu} (right) infers the governing preference from what the user has done before, verbalizes it (\eg, ``wash fruit before eating''), and plans on that basis. The verbalized preference is the intermediate representation bridging perception and action.}
    \label{fig:env}
\end{figure*}

Natural language is a flexible interface for human--robot collaboration, but the instructions it carries are rarely complete. By the Principle of Least Effort \citep{zipf2016human} and Gricean pragmatics \citep{grice1975logic}, speakers omit whatever they take the context to supply. In embodied settings the omission is consequential: a single instruction admits many valid executions, and which one is correct depends on both the environment and the user \citep{shridhar2020alfred,szot2021habitat,qiu2022emergent,wu2023tidybot}. ``Clean the kitchen'' fixes no answer to where items belong, what must be hand-washed, or in what order the subtasks proceed, and individuals answer these differently \citep{shridhar2020alfred,wu2023tidybot} (\cref{fig:env}). Personalized embodied agents must therefore move beyond literal instruction following toward \textit{pragmatic grounding}: recovering implicit preferences from prior behavior and using them to settle what later instructions leave open.

Foundation models have advanced multimodal reasoning and high-level planning in embodied environments, letting agents translate visual observations and natural-language instructions into grounded action plans \citep{brohan2023can,huang2023inner,driess2023palm,wang2023voyager,yang2025embodiedbench}. This progress has been measured almost entirely against explicit task specifications, and personalization has lagged behind: agents fall back on generic behavioral priors and miss what is idiosyncratic to a user \citep{zhang2024building}. Closing the gap requires treating preference as a latent variable that differs across users and inferring it from what they do, a problem close to \ac{tom} reasoning \citep{fawcett2010children,yuan2022situ,ying2026siftom}. Psychology suggests why behavior is the evidence to use: preferences are seldom articulated outright, and are instead constructed or revealed in the act of choosing \citep{epstein1994integration,slovic1995construction,lichtenstein2006construction,simonson2008will,zhu2020dark}. Embodied environments make the inference harder still, since the behavioral record available is multimodal, noisy, and few-shot.

The prevailing approach, \textit{implicit visual imitation}, conditions action generation directly on demonstration videos. In the few-shot regime this proves brittle. Demonstrations are high-dimensional and dominated by task-irrelevant variation in viewpoint, lighting, object instances, and scene layout, whereas the preference to be recovered is sparse and semantic (\eg, ``wash fruits before cutting,'' ``store perishables in the fridge''). Mapping pixels to actions in a single step asks the agent to identify the user-specific constraint and ground a plan at once, so that a failure of either becomes indistinguishable from a failure of the other. Scarce and costly demonstrations sharpen the difficulty \citep{akgun2012keyframe}. Preference-aware embodied learning has progressed in constrained domains such as rearrangement, yet remains task-specific and transfers poorly across the range of daily activities \citep{kapelyukh2022my,yuan2023learning,liu2025vlp}.

We separate the two stages with \ac{intu}, a simple two-stage formulation for few-shot personalized embodied planning that first verbalizes the latent user preference from demonstrations and then plans conditioned on that statement. The factorization serves two ends: it localizes failure to acquisition or to planning, and it tests whether an explicit semantic representation improves personalization at all. Compressing noisy multimodal demonstrations into a compact, human-interpretable rule (\eg, ``Preference: peel apples before serving'') interposes a semantic abstraction between what was observed and what is done. That explicit intermediate reasoning aids complex decision-making \citep{wei2022chain,wang2023plan}; we examine language as a representation of a user's preference.

Studying this systematically calls for a testbed in which the preference, rather than the task, is what varies. We build \ac{pbp}, a benchmark for \textit{pragmatic inference} in embodied planning, on a hierarchical taxonomy of 290 semantic preferences: Atomic Semantics fix fine-grained action parameters, Strategic Semantics fix categorical interaction and placement rules, and Temporal Semantics fix ordering and causal dependencies among subtasks. ALFRED \citep{shridhar2020alfred}, Habitat \citep{szot2021habitat}, and LoTa-Bench \citep{choi2024lota} score task completion against explicit specifications; \ac{pbp} instead asks whether an unspoken user-specific constraint can be recovered from few-shot demonstrations and carried into a new scene.

Three conditions expose a clear \textit{preference acquisition gap}: end-to-end planning from demonstrations, planning on preferences inferred through \ac{intu}, and planning on ground-truth preferences. Across \ac{sota} \acp{vlm} and \acp{llm}, supplying the preference outright yields markedly better plans than requiring the model to infer that same preference from behavior, which places the bottleneck in acquisition rather than in downstream planning. Verbalization narrows the gap: conditioning on an inferred linguistic preference improves alignment over end-to-end conditioning throughout. Performance rises with additional demonstrations and holds up better when preferences must transfer across visually distinct scenes. Explicit semantic representations thus supply a serviceable abstraction for carrying sparse user-specific constraints from behavior into action, leaving visual-to-semantic preference inference as the principal obstacle for current models.

Our contributions are threefold:
\textbf{(i) \ac{pbp} benchmark:} a large-scale benchmark that shifts evaluation from explicit instruction following to implicit preference inference, over 290 preferences organized by a hierarchical semantic taxonomy;
\textbf{(ii) Preference acquisition gap:} a systematic diagnosis showing that current embodied agents plan effectively once user preferences are specified, but degrade substantially when the same preferences must be inferred from few-shot behavioral demonstrations;
\textbf{(iii) Language-mediated preference acquisition:} evidence, through \ac{intu}, that verbalizing latent preferences before planning narrows this gap, and that language can therefore serve as an effective semantic abstraction for few-shot embodied personalization.

\section{Related Work}

\paragraph{Instruction following and pragmatic grounding}
Language-guided embodied agents have traditionally executed explicit task specifications, from \ac{vln} \citep{anderson2018vision,chen2019touchdown} to household manipulation and rearrangement \citep{shridhar2020alfred,szot2021habitat}. Human instructions, however, are rarely exhaustive: speakers omit what shared context can recover, as the principle of least effort \citep{zipf2016human} and pragmatic theories of communication \citep{grice1975logic} would predict. Embodied systems have accordingly begun reasoning about implicit goals, contextual cues, and human mental states \citep{jiang2022what,sarch2022tidee,xu2023active,ying2026siftom}. Yet these formulations resolve ambiguity from environmental context or generic commonsense, not from behavioral evidence about the particular user. We study the complementary problem of \textit{pragmatic grounding from user history}: recovering latent user-specific preferences from a handful of behavioral demonstrations and applying them to underspecified instructions in new scenes.

\paragraph{Language as an intermediate representation}
Language increasingly serves as an intermediate representation for reasoning and planning. Chain-of-Thought and related prompting show that externalizing intermediate reasoning improves complex language reasoning \citep{wei2022chain,wang2023plan}, while language and code supply structured abstractions between perception and action for task decomposition, affordance reasoning, and long-horizon planning \citep{brohan2023can,liang2023code,wang2023voyager,xu2026code,liu2025survey}. Recasting multimodal observations as structured text or symbols likewise simplifies downstream planning by separating perceptual interpretation from action generation. We investigate a different intermediate variable: \textit{user preference}. Rather than decomposing a task or describing an environment, we verbalize the latent behavioral regularity inferred from demonstrations before planning, and use this factorization as a diagnostic intervention: whether explicit preference acquisition accounts for the gap between observing user behavior and generating personalized plans.

\paragraph{Personalization and theory of mind in agents}
Personalization is well studied in recommendation and dialogue, and recent work extends it to \acp{llm} through user-specific preferences, memories, and interaction histories \citep{packer2023memgpt,memoryllm,xu2025cross}. Embodied personalization adds a further demand: what is specific to a user must be inferred from situated interaction and behavior. Early robotic systems addressed preferences in constrained domains such as object rearrangement and household organization \citep{kapelyukh2022my,wu2023tidybot}, and recent personalized embodied agents pursue memory-based user modeling and personalized context accumulation \citep{ziliotto2026personal,kwon2026embodied,lee2026personalizing}. These lines of work concern how user-specific information is retrieved or represented; \ac{pbp} isolates a prior question, whether such preferences can be acquired from behavioral demonstrations at all and then used to resolve underspecified instructions. Separating acquisition from preference-conditioned planning gives a controlled account of where personalized embodied agents fail.

\section{Problem Formulation}

We cast personalized embodied planning as context-dependent preference inference: the agent must draw on prior behavioral evidence to resolve what the current instruction leaves open.

\subsection{Task definition}

Consider an embodied agent in a physical environment $\mathcal{E}$. The agent receives a natural-language instruction $I$ (\eg, ``Prepare an apple'') that is \textbf{semantically underspecified} and admits multiple valid executions. To resolve the ambiguity, it is given a few-shot behavioral context
\begin{equation}
    \mathcal{C}_{\mathrm{demo}} = \{X_i\}_{i=1}^{K},
\end{equation}
in which each $X_i$ is an unlabeled demonstration, rendered as an egocentric video or, in the symbolic setting, as the corresponding action trace.

Demonstrations vary in behavioral pattern and task configuration, and each may exhibit several regularities at once, so no single demonstration determines the relevant preference on its own. What makes a regularity relevant is the query: given $\mathcal{C}_{\mathrm{demo}}$, the instruction $I$, and the current environment $\mathcal{E}_{\mathrm{curr}}$, the agent must recover the latent preference bearing on the task at hand,
\begin{equation}
    \hat{S}_{\mathrm{pref}}
    =
    \arg\max_S
    P(S \mid \mathcal{C}_{\mathrm{demo}}, I, \mathcal{E}_{\mathrm{curr}}),
\end{equation}
and then act on it,
\begin{equation}
    \hat{A}
    =
    \arg\max_A
    P(A \mid I, \hat{S}_{\mathrm{pref}}, \mathcal{E}_{\mathrm{curr}}).
\end{equation}

\subsection{Language-mediated preference acquisition}

Planning end-to-end from demonstrations entangles two distinct problems: identifying which preference governs the current task, and grounding an action sequence that honors it. \ac{intu} separates them by making the linguistic representation $S_{\mathrm{pref}}$ an explicit intermediate variable. Instead of mapping behavioral history straight to actions, \ac{intu} reasons over the demonstrations in light of the current instruction and environment, states the inferred preference as a semantic constraint, and conditions planning on that statement. The two stages can then be scored apart: whether a model recovers the right preference from behavior, and whether it plans well once the preference is handed to it.

\section{The \texorpdfstring{\ac{pbp}}{PbP} Benchmark}

\begin{figure}[t!]
    \centering
    \includegraphics[width=\linewidth]{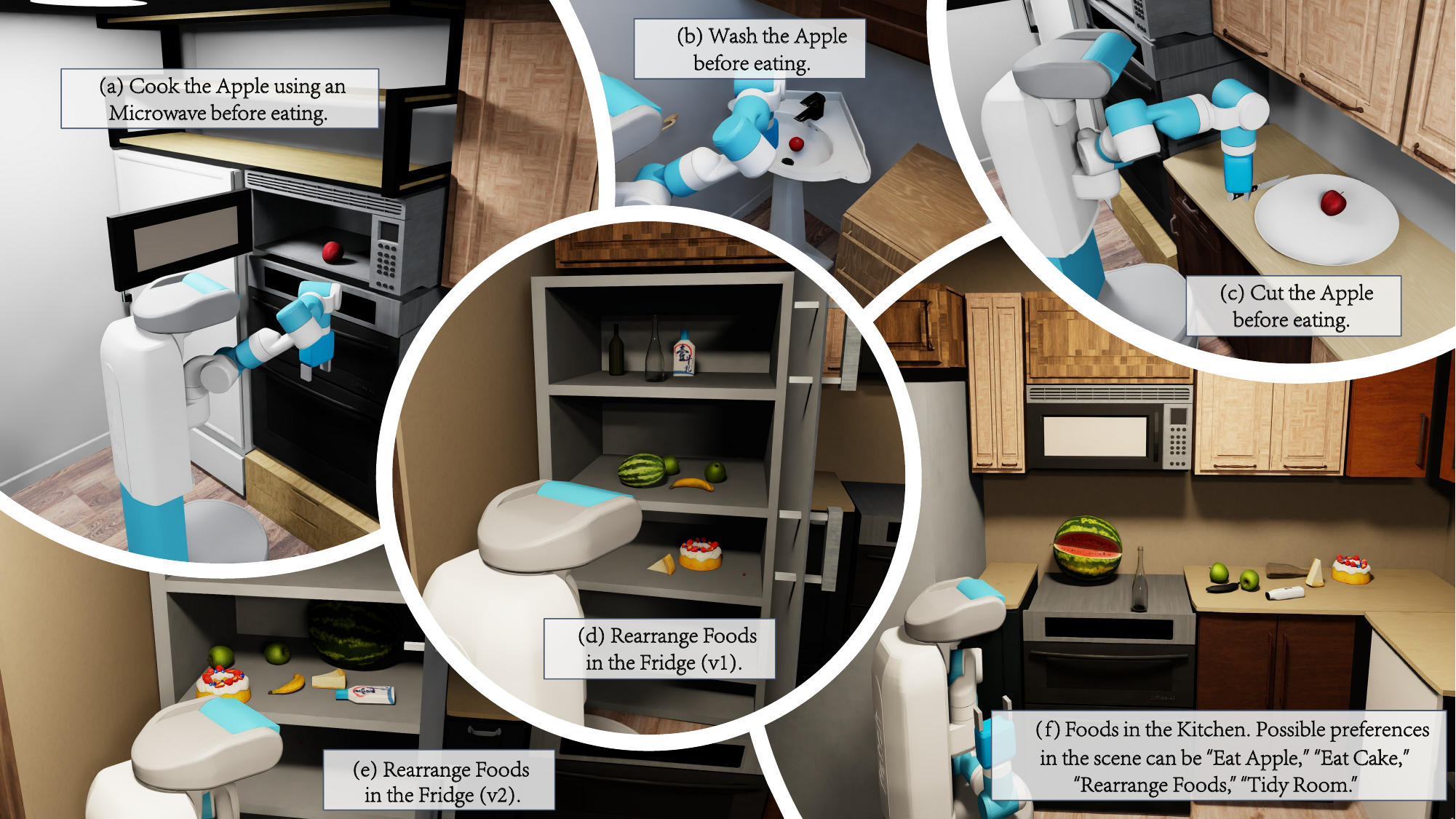}
    \caption{\textbf{Preference levels in \acs{pbp}.} The three pragmatic levels differ in what must be abstracted from behavior. (a--c) Atomic Semantics turn on a primitive action parameter, here whether the apple is microwaved, washed, or cut before eating. (d--e) Strategic Semantics turn on a categorical rule, here two incompatible schemes for arranging the same foods in a fridge. (f) Temporal Semantics turn on task order, which a single scene leaves ambiguous: the same kitchen affords eating an apple, eating cake, rearranging foods, or tidying, and only prior behavior settles which comes first.}
    \label{fig:dataset}
\end{figure}

\begin{figure*}[t!]
    \centering
    \small
    \includegraphics[width=\linewidth]{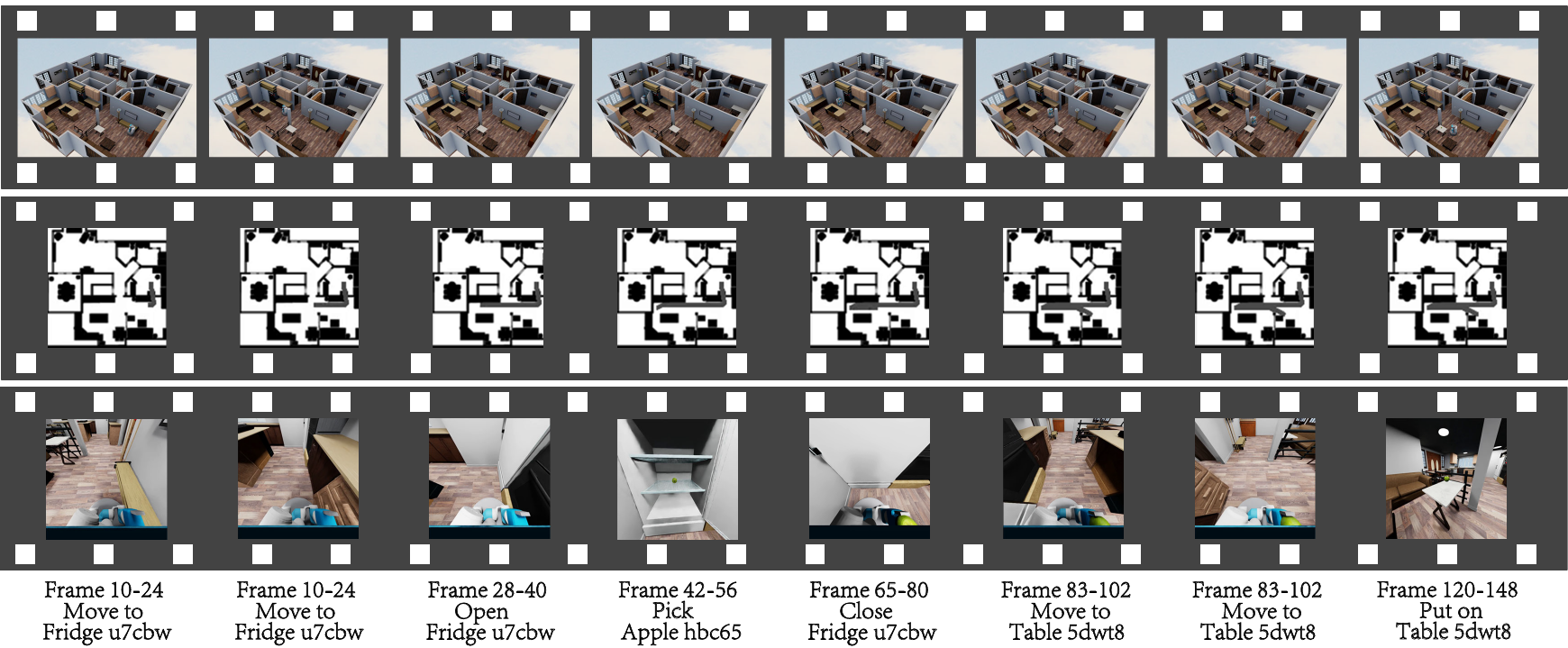}
    \caption{\textbf{Multimodal demonstration data in \acs{pbp}.} Each demonstration supplies part of the context $\mathcal{C}_{\mathrm{demo}}$ from which the latent constraint $S_{\mathrm{pref}}$ must be inferred (\eg, ``place apple on table''). The agent observes the egocentric stream (bottom) together with a bird's-eye map (middle) for spatial context; the third-person view (top) is rendered for visualization only. Frame-level action labels mark the primitives composing the episode.}
    \label{fig:datapoint}
\end{figure*}

We introduce \ac{pbp}, a benchmark for pragmatic inference in embodied planning. Prior embodied benchmarks score navigation or the execution of explicit manipulation goals \citep{shridhar2020alfred}; \ac{pbp} instead requires that a latent semantic constraint be inferred from few-shot demonstrations and carried into a new, underspecified instruction.

\subsection{Semantic taxonomy of preferences}

\ac{pbp} organizes 290 preferences into three pragmatic levels of increasing abstraction (\cref{fig:dataset}). \textbf{Atomic Semantics (action level)} fix execution parameters that natural instructions leave out (\eg, ``fill halfway'' \vs\ ``fill completely''), so the difficulty lies in grounding subtle continuous state differences as discrete semantic attributes. \textbf{Strategic Semantics (option level)} fix interaction and placement policies (\eg, ``store perishables in the fridge''), which demands category-level abstraction: mapping instances onto abstract properties and carrying the rule across scenes. \textbf{Temporal Semantics (sequence level)} fix ordering and causal dependencies among subtasks (\eg, ``clean surfaces before placing items''), where the preference resides in relative action order rather than in any individual action. \cref{sec:supp:dataset_statis} details the dataset and the full hierarchy.

\subsection{Data construction}

\ac{pbp} is built in OmniGibson and comprises 5,000 evaluation groups. Each group pairs (i) a behavioral history of $K{=}3$ unlabeled demonstration episodes spanning several tasks and scenes with (ii) a query consisting of a new scene and an underspecified instruction. Since the history encodes more than one regularity, the query is what selects the preference at issue. Gold annotations accompany every episode, including preference metadata and a canonical action sequence; neither is exposed to the model in the main preference-inference setting.

\subsection{Dataset statistics and diversity}

\ac{pbp} spans 50 interactive OmniGibson scenes (kitchens, living rooms, offices, grocery stores) and over 3,000 object assets, so that a preference cannot be recognized by appearance alone. The 290 preferences divide as:
\begin{itemize}[nosep,leftmargin=*]
    \item Action level: 75 types (continuous parameters, \eg, tap temperature, slicing thickness);
    \item Option level: 135 types (storage and interaction strategies, \eg, cabinet \vs\ counter);
    \item Sequence level: 80 types (long-horizon ordering constraints over 3 or more subtasks).
\end{itemize}
Together the taxonomy and the episode structure stress fine-grained perception and high-level semantic generalization at once. The main experiments concentrate on the option and sequence levels, whose categorical and temporal demands cannot be met by direct behavioral imitation, making them the sharper test of language-mediated preference acquisition; action-level preferences are included in the benchmark for completeness.

\section{Experiments}

\begin{table*}[t!]
    \small
    \centering
    \setlength{\tabcolsep}{3pt}
    \caption{\textbf{Planning error under three sources of preference} (Levenshtein distance $\downarrow$; lower is better aligned with the user). \colorbox{LightCyan1}{End-to-end} plans directly from demonstrations and aligns poorly. \colorbox{MistyRose1}{Two-stage} verbalizes the inferred preference first and cuts the error roughly in half. \colorbox{MistyRose1}{Second-stage (gt)} substitutes the ground-truth preference, so its residual is planning error alone; the distance between it and \colorbox{MistyRose1}{Two-stage} is what acquisition costs. \acs{vivit} has no language stage and appears only as an end-to-end lower bound. Values are mean $\pm$ range over 3 independent runs.}%
    \label{tab:results_levenshtein_distance}%
    \resizebox{\textwidth}{!}{%
        \begin{tabular}{cccccccccc}
            \toprule
            & \multicolumn{4}{c}{\textbf{VIDEO-BASED INPUT}} & & \multicolumn{4}{c}{\textbf{SYMBOL-BASED INPUT}}\\
            \midrule
            & ViViT & LLaVA-NeXT & EILEV & Gemini-3.1-Pro & & Llama3-8B & DS-v4-pro & Claude-Opus-4.7 & GPT-5.5 \\
            \multicolumn{10}{c}{\textbf{Option Level}}\\
            \cellcolor{LightCyan1} \textbf{End-to-end} & \cellcolor{LightCyan1} 15.49$\pm$1.29 & \cellcolor{LightCyan1} 15.94$\pm$3.41 & \cellcolor{LightCyan1} 12.88$\pm$2.20 & \cellcolor{LightCyan1} 8.72$\pm$2.26  & \cellcolor{LightCyan1} & \cellcolor{LightCyan1} 14.74$\pm$3.21 & \cellcolor{LightCyan1} 6.42$\pm$2.33 & \cellcolor{LightCyan1} 6.72$\pm$2.48 & \cellcolor{LightCyan1} 6.07$\pm$2.00  \\
            \cellcolor{MistyRose1} \textbf{Two-stage} & \cellcolor{MistyRose1} - & \cellcolor{MistyRose1} 12.46$\pm$3.23 & \cellcolor{MistyRose1} 12.89$\pm$3.74 & \cellcolor{MistyRose1} 4.28$\pm$1.76 & \cellcolor{MistyRose1} & \cellcolor{MistyRose1} 9.67$\pm$5.16 & \cellcolor{MistyRose1} 2.31$\pm$1.67 & \cellcolor{MistyRose1} 2.56$\pm$2.04 & \cellcolor{MistyRose1} 2.41$\pm$1.72 \\
            \cellcolor{MistyRose1} \textbf{Second-stage (gt)} & \cellcolor{MistyRose1} - & \cellcolor{MistyRose1} 3.28$\pm$5.29 & \cellcolor{MistyRose1} 11.18$\pm$4.20 & \cellcolor{MistyRose1} 0.16$\pm$1.82 & \cellcolor{MistyRose1} & \cellcolor{MistyRose1} 8.22$\pm$5.58 & \cellcolor{MistyRose1} 0.16$\pm$1.46 & \cellcolor{MistyRose1} 0.16$\pm$1.99 & \cellcolor{MistyRose1} 0.14$\pm$1.21\\
            \midrule
            \multicolumn{10}{c}{\textbf{Sequence Level}}\\
            \cellcolor{LightCyan1} \textbf{End-to-end} & \cellcolor{LightCyan1} 34.04$\pm$11.84 & \cellcolor{LightCyan1} 34.76$\pm$11.25 & \cellcolor{LightCyan1} 33.10$\pm$12.21 & \cellcolor{LightCyan1} 24.51$\pm$4.15 & \cellcolor{LightCyan1} & \cellcolor{LightCyan1} 31.79$\pm$7.32 & \cellcolor{LightCyan1} 22.84$\pm$4.01 & \cellcolor{LightCyan1} 21.47$\pm$3.12 & \cellcolor{LightCyan1} 21.92$\pm$3.45 \\
            \cellcolor{MistyRose1} \textbf{Two-stage} & \cellcolor{MistyRose1} - & \cellcolor{MistyRose1} 30.02$\pm$13.54 & \cellcolor{MistyRose1} 33.03$\pm$13.61 & \cellcolor{MistyRose1} 12.57$\pm$2.92 & \cellcolor{MistyRose1} & \cellcolor{MistyRose1} 25.46$\pm$5.93 & \cellcolor{MistyRose1} 12.10$\pm$2.85 & \cellcolor{MistyRose1} 13.84$\pm$2.81 & \cellcolor{MistyRose1} 12.48$\pm$2.45 \\
            \cellcolor{MistyRose1} \textbf{Second-stage (gt)} & \cellcolor{MistyRose1} - & \cellcolor{MistyRose1} 18.92$\pm$14.18 & \cellcolor{MistyRose1} 26.57$\pm$12.21 & \cellcolor{MistyRose1} 10.24$\pm$2.58 & \cellcolor{MistyRose1} & \cellcolor{MistyRose1} 19.02$\pm$7.10 & \cellcolor{MistyRose1} 10.19$\pm$2.47 & \cellcolor{MistyRose1} 9.17$\pm$2.90 & \cellcolor{MistyRose1} 9.77$\pm$2.69 \\
            \midrule
            \multicolumn{10}{c}{\textbf{Overall}}\\
            \cellcolor{LightCyan1} \textbf{End-to-end} & \cellcolor{LightCyan1} 24.76 & \cellcolor{LightCyan1} 25.35 & \cellcolor{LightCyan1} \textbf{22.99} & \cellcolor{LightCyan1} 16.62 & \cellcolor{LightCyan1} & \cellcolor{LightCyan1} 23.26 & \cellcolor{LightCyan1} 14.63 & \cellcolor{LightCyan1} 14.10 & \cellcolor{LightCyan1} 14.00 \\
            \cellcolor{MistyRose1} \textbf{Two-stage} & \cellcolor{MistyRose1} - & \cellcolor{MistyRose1} 21.24 & \cellcolor{MistyRose1} 22.96 & \cellcolor{MistyRose1} 8.43 & \cellcolor{MistyRose1} & \cellcolor{MistyRose1} 17.56 & \cellcolor{MistyRose1} 7.21 & \cellcolor{MistyRose1} 8.20 & \cellcolor{MistyRose1} 7.45 \\
            \cellcolor{MistyRose1} \textbf{Second-stage (gt)} & \cellcolor{MistyRose1} - & \cellcolor{MistyRose1} 11.10 & \cellcolor{MistyRose1} 18.88 & \cellcolor{MistyRose1} 5.20 & \cellcolor{MistyRose1} & \cellcolor{MistyRose1} 13.62 & \cellcolor{MistyRose1} 5.18 & \cellcolor{MistyRose1} 4.67 & \cellcolor{MistyRose1} 4.96 \\
            \bottomrule
        \end{tabular}%
    }%
\end{table*}

Our experiments ask whether explicit preference acquisition supports robust personalized embodied planning, along four lines. First, whether end-to-end failures originate in planning itself or in the absence of a preference representation. Second, what verbalizing the preference buys over conditioning on demonstrations implicitly. Third, how acquisition and planning scale with the number of demonstrations. Fourth, whether an inferred preference survives transfer to a visually distinct scene.

\subsection{Experimental setup}

\paragraph{Models}
Video-based inference is tested with Gemini-3.1-Pro, \acs{llava}-NeXT \citep{liu2024improved}, and \acs{EILEV} \citep{yu2023efficient}. Gemini-3.1-Pro, a closed-source \ac{vlm}, receives keyframes extracted at 1\,Hz and interleaved with text prompts. \acs{llava}-NeXT (Video) encodes the demonstrations natively as video. \acs{EILEV}, built for few-shot in-context learning from embodied video, tests whether specialized pre-training helps preference inference. We add \ac{vivit} \citep{arnab2021vivit}, a pure-Transformer video encoder trained end-to-end on spatio-temporal features; lacking any \ac{llm} component, it lower-bounds commonsense-driven pragmatic inference in \ac{pbp}. To disentangle reasoning from perception we also evaluate \acp{llm} on symbolic input (GPT-5.5, Claude-Opus-4.7, DeepSeek-v4-pro, Llama-3-8B), replacing the visual demonstrations with ground-truth scene graphs and action logs. GPT-5.5 and Claude-Opus-4.7 accept video as well; the symbolic condition is used solely to isolate reasoning under perfect perception and thus approximates an upper bound on the planning side. We further provide a symbolic causal discovery model DAG-Opt \citep{zheng2018dags}.

\paragraph{Protocols}
The \textbf{End-to-End} setting has the model generate an action sequence directly from the instruction and demonstrations, with no mention of latent preferences in the prompt, as users rarely articulate every constraint in practice. The \textbf{Language-Mediated} setting has it first verbalize an intermediate preference inferred from the demonstrations, then plan conditioned on that statement. Between these sits a \textbf{Preference-aware End-to-End} baseline, which asks the model to infer the preference internally but emit only the plan; comparing the three separates the benefit of reasoning about preferences from the benefit of exposing the result. Prompts follow the OpenAI Cookbook\footnote{\url{https://cookbook.openai.com/}}, and every experiment is run three times with fixed hyperparameters and no seed tuning. \Cref{sec:supp:baseline} gives implementation details and full prompts.

\paragraph{Isolating preference from comprehension}
A rule-based diagnostic confirms that every evaluated \ac{llm} satisfies the literal instruction over 95\% of the time, where literal satisfaction asks only whether the plan addresses the stated objective, such as manipulating the requested object, irrespective of any hidden preference. The alignment gap we report therefore reflects preference inference and not a failure to understand the instruction.

\begin{table*}[t!]
    \small
    \centering
    \caption{\textbf{Accuracy of the verbalized preference} ($\uparrow$), measuring whether $S_{\mathrm{pref}}$ is named correctly before any planning occurs. \textbf{Few-shot} supplies the $K{=}3$ demonstrations; \textbf{Ablative} withholds them, leaving only the instruction and scene. The collapse under Ablative, sharpest at the sequence level, shows that accuracy rests on the user's demonstrated behavior rather than on commonsense defaults about where things belong.}%
    \label{tab:results_combined}%
    \rowcolors{2}{LightCyan1}{}
    \resizebox{\linewidth}{!}{%
        \begin{tabular}{cccccccccccc}
            \toprule
            & \multicolumn{4}{c}{\textbf{VIDEO-BASED INPUT}} & & \multicolumn{5}{c}{\textbf{SYMBOL-BASED INPUT}}\\
            \midrule
            
            & ViViT & LLaVA-NeXT & EILEV & Gemini-3.1-Pro & & DAG-Opt & Llama3-8B & DS-v4-pro & Claude-Opus-4.7 & GPT-5.5 \\
            \multicolumn{11}{c}{\textbf{Few-shot (With Context Demonstrations)}}\\
            \textbf{Option Level} & 9.38 & 36.87 & 38.33 & 63.28 & & 10.15 & 72.98  & 91.28 & 89.91 & 92.48 \\
            \textbf{Sequence Level} & 4.24 & 24.85 & 32.69 & 52.10 & & 13.49 & 67.18  & 76.48 & 76.92 & 78.01 \\
            \textbf{Overall} & 6.81 & 30.86 & 35.51 & 57.69 & & 11.82 & 70.08  & 83.88 & 83.42 & 85.25 \\
            \midrule
            \multicolumn{11}{c}{\textbf{Ablative (Zero-shot / No Context)}}\\
             \textbf{Option Level} & 9.16 & 15.47 & 4.77 & 31.92 & & 3.84 & 39.50  & 83.55 & 82.47 & 82.01 \\
            \textbf{Sequence Level} & 4.38 & 8.13 & 0.00 & 14.28 & & 1.28 & 6.25  & 18.58 & 18.02  & 18.71 \\
            \textbf{Overall} & 6.77 & 11.80 & 2.38 & 23.10 & & 2.56 & 22.88 & 51.07 & 50.25 & 50.36 \\
            \bottomrule
        \end{tabular}%
    }%
\end{table*}

\subsection{Evaluation metrics}

Two metrics track the two stages of the pipeline.

\paragraph{Preference inference accuracy (Acc).} Whether the verbalized preference $\hat{S}_{\mathrm{pref}}$ semantically matches the ground truth, judged by strict keyword matching together with semantic similarity checks.

\paragraph{Action alignment (Levenshtein distance).} The edit distance between the generated sequence $\hat{A}$ and the canonical sequence $A_{\mathrm{gt}}$, counting insertions, deletions, and substitutions. Being graded rather than binary, it discriminates degrees of misalignment that a success rate would collapse; lower is better. Where objects are interchangeable, an order-invariant mask keeps permutations over equivalents from being penalized. Strict ordering is enforced only for sequence-level preferences, in which the order is itself the preference.

\subsection{Results and analysis}

\noindent \textbf{RQ1: Where does the error originate?}
\Cref{tab:results_levenshtein_distance} varies how the preference reaches the planner while holding the planner itself fixed. Handed the ground-truth preference (\textit{Second-stage (gt)}), every model plans markedly better, and at the option level the strongest reach essentially zero edit distance. Sequence-level error persists, so specifying the preference removes much of the total error without dissolving the difficulty of long-horizon planning. Since nothing about the downstream architecture changed between conditions, the difference is attributable to acquisition. The modality comparison points the same way: Gemini-3.1-Pro closes much of the video--symbol gap left by earlier \acp{vlm} (overall end-to-end 16.62 \vs\ 24.76 for \acs{vivit}), yet still trails the symbolic models (GPT-5.5, 14.00), which locates the residual difficulty in visual-to-semantic acquisition rather than in planning.

\noindent \textbf{RQ2: What does verbalization contribute?}
Across \cref{tab:results_levenshtein_distance}, stating the preference before planning lowers error consistently. Requiring a discrete constraint (\eg, ``the user prefers option A'') relieves the planner of recovering intent and generating actions at once, and it renders acquisition errors visible rather than leaving them to surface as plan defects. \Cref{tab:preference_aware_baseline} apportions the credit: the preference-aware end-to-end baseline already improves substantially over vanilla end-to-end, so reasoning about the latent preference is what matters most. Two-stage verbalization adds a further, consistent gain, largest at the sequence level, and exposes the intermediate representation for inspection and reuse.

\noindent \textbf{RQ3: How much does context matter?}
\textit{Preference prediction.} \Cref{tab:results_combined} isolates the first stage, where inference proves far from solved. Video-based models trail their symbolic counterparts throughout (Gemini-3.1-Pro, 57.69\%, against GPT-5.5 at 85.25\% and DeepSeek-v4-pro at 83.88\%). The stratification is by input modality, not by model strength, which again identifies visual-to-text alignment as the binding constraint.

\textit{Few-shot \vs\ zero-shot.} Withholding the demonstrations collapses accuracy across the board (\cref{tab:results_combined}, bottom): GPT-5.5 falls from 85.25\% to 50.36\%, Gemini-3.1-Pro from 57.69\% to 23.10\%. What survives is instructive. Option-level accuracy remains respectable because commonsense supplies a default, yet defaults do not settle which of several sound alternatives a particular user favors, eggs in the fridge or in the pantry. Sequence-level accuracy all but vanishes, temporal preferences being the least recoverable from generic priors.

\begin{figure*}[t!]
    \centering
    \small
    \begin{subfigure}{0.49\linewidth}
        \centering
        \includegraphics[width=0.99\linewidth]{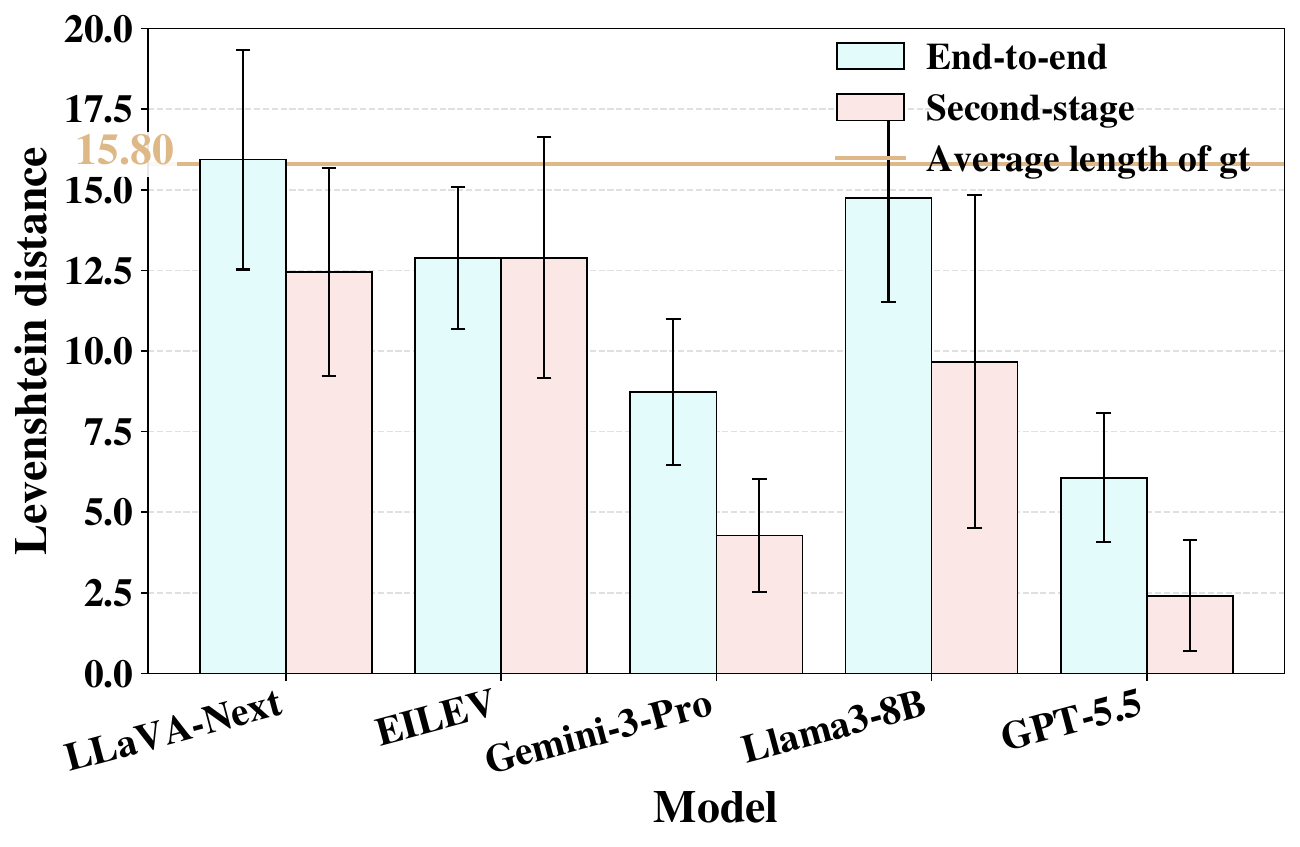}
        \caption{Option level}
    \end{subfigure}%
    \begin{subfigure}{0.49\linewidth}
        \centering
        \includegraphics[width=0.99\linewidth]{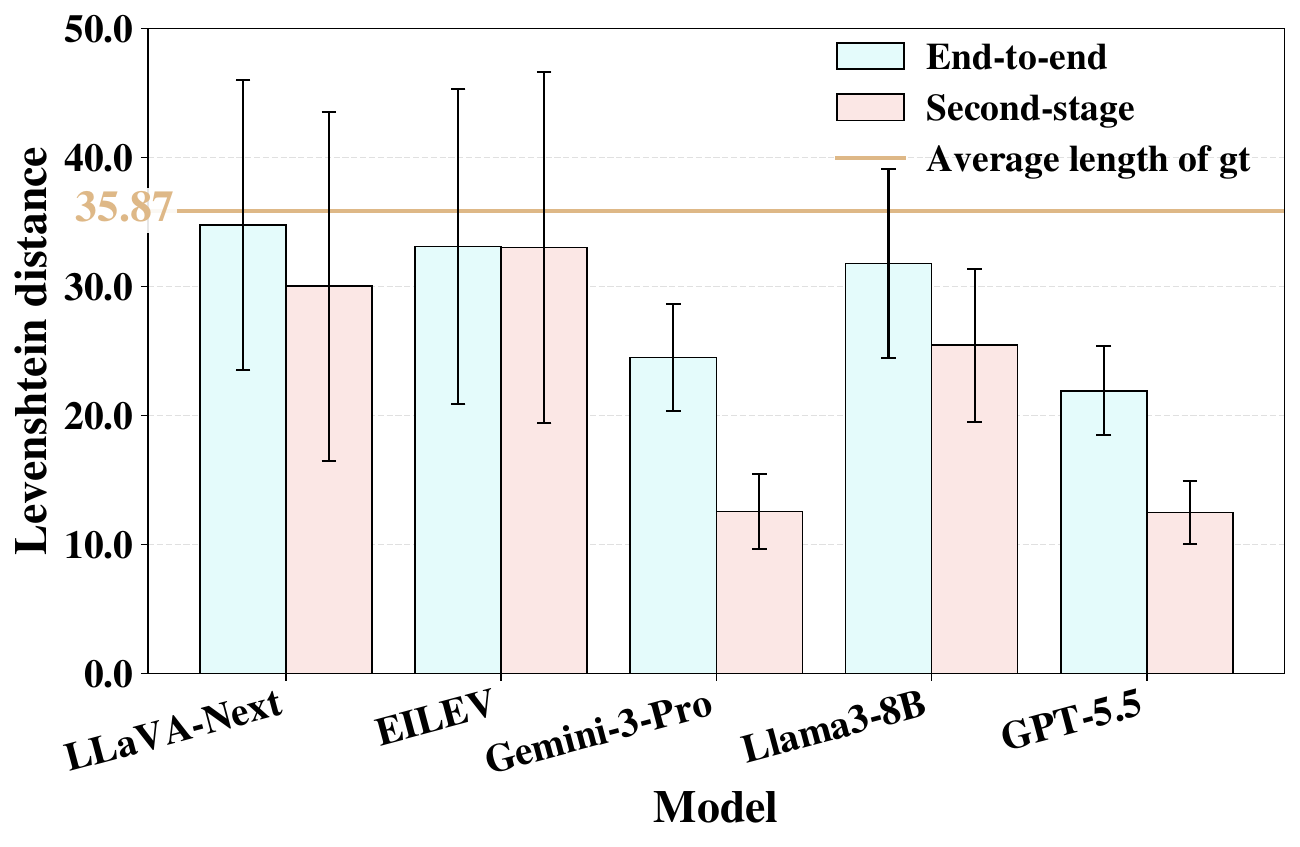}
        \caption{Sequence level}
    \end{subfigure}%
    \\%
    \begin{subfigure}{0.49\linewidth}
        \centering
        \includegraphics[width=0.99\linewidth]{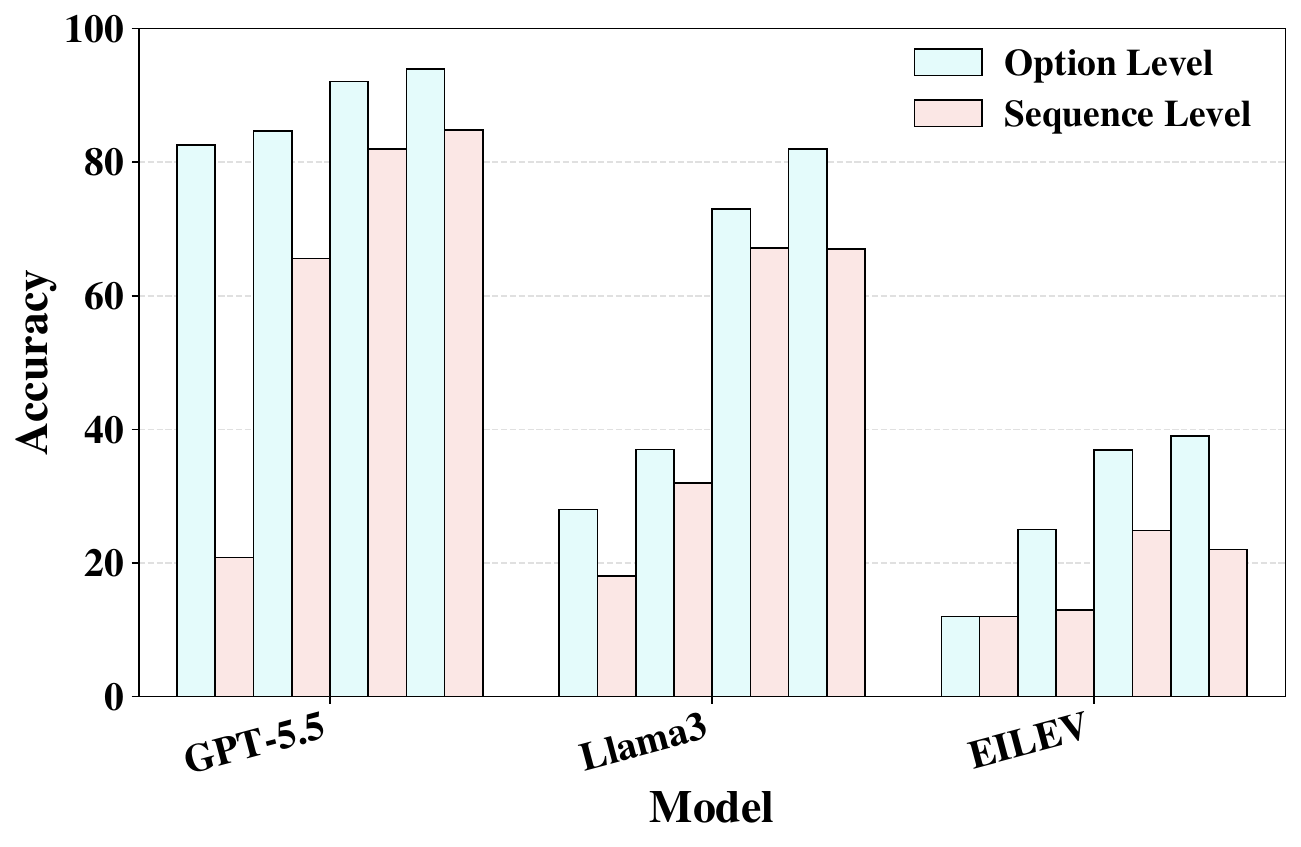}
        \caption{Preference learning}
    \end{subfigure}%
    \begin{subfigure}{0.49\linewidth}
        \centering
        \includegraphics[width=0.99\linewidth]{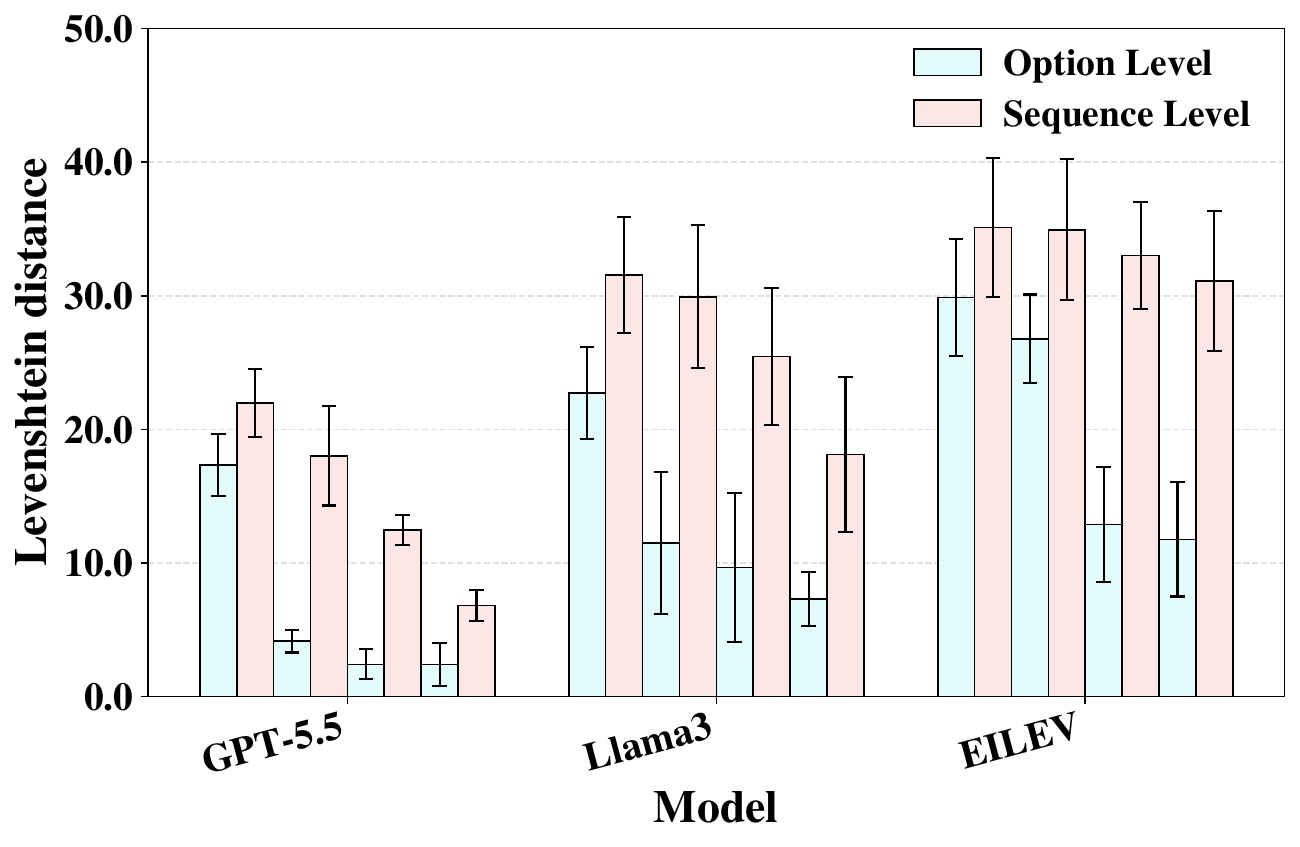}
        \caption{Action planning}
    \end{subfigure}%
    \caption{\textbf{Effect of verbalization and of context size.} (a--b) Verbalization narrows the alignment gap at both the (a) option and (b) sequence level: \colorbox{LightCyan1}{End-to-end} generates the sequence directly from prior observations, whereas \colorbox{MistyRose1}{Two-stage} first receives a predicted preference label. The \textcolor{BurlyWood}{---} line marks the average ground-truth sequence length (15.80 for option level, 35.87 for sequence level), against which the residual distances should be read. (c--d) Enlarging the support set ($k \in \{1,2,3,5\}$) generally improves both stages, (c) preference learning and (d) action planning, on \colorbox{LightCyan1}{Option Level} and \colorbox{MistyRose1}{Sequence Level} tasks alike, though the gain saturates in the visual encoders. Higher is better in (c), lower in (d).}
    \label{fig:distance}
\end{figure*}

\textit{Scaling demonstrations.} \Cref{fig:distance}(c--d) tracks support sets of $k \in \{1,2,3,5\}$. Additional demonstrations improve both prediction accuracy and downstream planning, but the returns diminish, and for \acs{EILEV} the largest context ($k{=}5$) occasionally hurts, its extra frames arriving as noise. More context is not uniformly better when the stream is long and multimodal.

\noindent \textit{Diagnostic takeaway.} The option level probes category abstraction, the sequence level temporal-causal reasoning, and the latter costs consistently $2\text{--}3\times$ more error (Gemini-3.1-Pro, 8.72 \vs\ 24.51). Abstracting temporal structure from visual streams is the sharpest remaining difficulty.

\begin{table}[t!]
    \small
    \centering
    \setlength{\tabcolsep}{3pt}
    \caption{\textbf{Preference inference across scene shift} (accuracy $\uparrow$). Under \textit{direct}, demonstration and test episodes share a room and its objects; under \textit{gen}, they share only the preference. The symbolic models hold most of their accuracy across the shift, whereas the vision-based models give up more of theirs, having partly bound the preference to the scene it was demonstrated in.}
    \label{table:results_generalization}
    \rowcolors{3}{LightCyan1}{}
    \begin{tabular}{lcccc}
        \toprule
        & \multicolumn{2}{c}{\textbf{Option Level}} & \multicolumn{2}{c}{\textbf{Sequence Level}} \\
        \cmidrule(lr){2-3} \cmidrule(lr){4-5}
        \textbf{Model} & \textit{direct} & \textit{gen} & \textit{direct} & \textit{gen} \\
        \midrule
        LLaVA-NeXT      & 33.25 & 36.87 & 33.12 & 24.85 \\
        EILEV           & 46.93 & 38.33 & 37.53 & 32.69 \\
        Gemini-3.1-Pro  & 65.89 & 63.28 & 68.21 & 52.10 \\
        DS-v4-pro       & 95.18 & 91.28 & 86.42 & 76.48 \\
        GPT-5.5         & 95.01 & 92.48 & 86.90 & 78.01 \\
        \bottomrule 
    \end{tabular}%
\end{table}

\noindent \textbf{RQ4: Does an inferred preference survive a change of scene?}
A preference worth learning must outlast the room it was learned in. We therefore contrast a \textit{direct} setting, in which demonstration and test episodes are rendered in the same room with the same objects, against a \textit{generalization} (\textit{gen}) setting, in which they share the preference but nothing of the scene. \Cref{table:results_generalization} separates the two families cleanly at the sequence level, where the ordering constraint cannot be read off any single frame. The symbolic models surrender little (GPT-5.5, 86.90 to 78.01; DeepSeek-v4-pro, 86.42 to 76.48): once a preference has been put into words it is invariant to pixels. The vision-based models give up more, Gemini-3.1-Pro falling from 68.21 to 52.10 and \acs{EILEV} from 37.53 to 32.69, which suggests their few-shot ``learning'' attaches partly to background texture and object layout rather than to the behavioral rule, so that changing the scene breaks the correlation. Option-level accuracy is more stable for every model, commonsense priors carrying some of the burden there; \acs{llava}-NeXT even scores marginally higher under \textit{gen}, though at roughly a third accuracy its two conditions are close to indistinguishable.

\Cref{fig:test_points} resolves this per sample. Certain scenes defeat all models alike, a property of the environment rather than of any architecture. More tellingly, the vision-based models succeed on largely disjoint samples under \textit{direct} and \textit{gen}: what they acquire holds in the room where it was shown and lapses elsewhere. Verbalization counteracts precisely this, forcing scene-specific detail to be dropped and only the transferable rule retained.

\subsection{Further Discussion}

\begin{figure*}[t!]
    \small
    \centering
    \includegraphics[width=\linewidth]{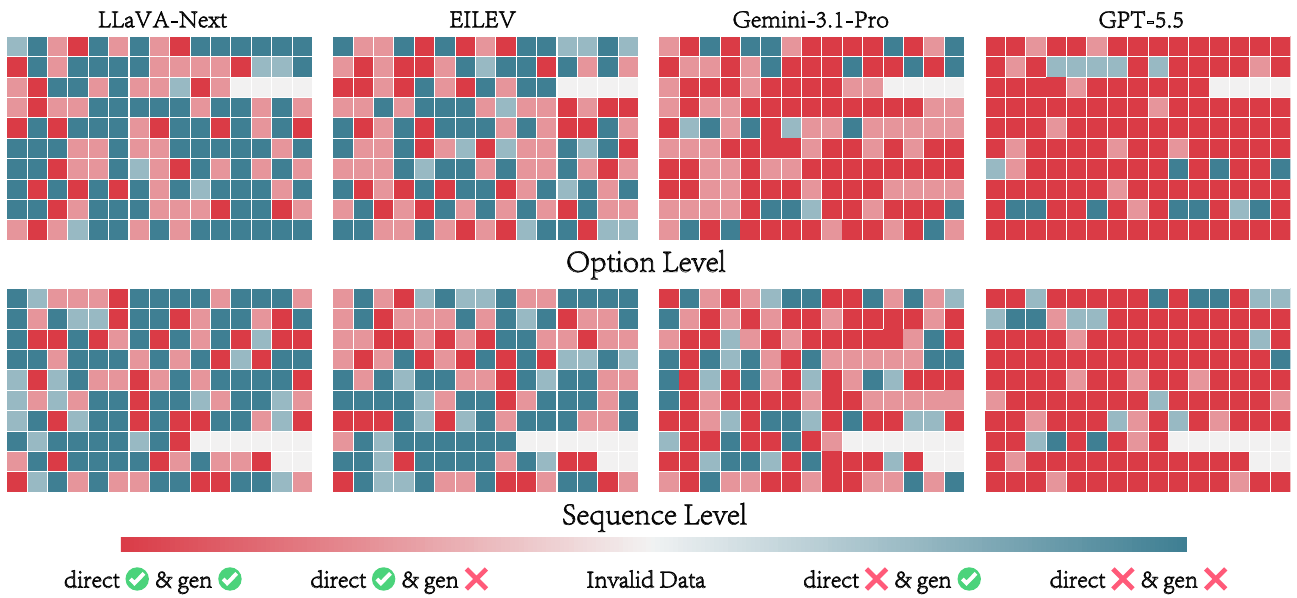}
    \caption{\textbf{Per-sample outcomes under \textit{direct} and \textit{gen}.} Each row is a distinct scene and each cell a test sample, colored by whether the model succeeds in both settings, in neither, or in only one. Two patterns recur: some scenes defeat every model, and the vision-based models succeed on largely disjoint samples across the two settings, a signature of scene-dependent rather than rule-based inference.}
    \label{fig:test_points}
\end{figure*}

Language serves here not as a channel for communication but as a compact semantic interface between behavioral evidence and planning. Raw demonstrations carry a great deal of perceptual variation that bears on nothing; the preference within them is sparse and semantic. Verbalization compresses the one into the other, keeping intent and shedding trajectory detail, and it is this compression that carries preferences across environments.

Recovering the preference in the first place remains hard. Gemini-3.1-Pro, among the strongest \acp{vlm} available, names only about 63\% of option-level and 52\% of sequence-level preferences correctly (\cref{tab:results_combined}), so abstracting latent intent from behavior is far from solved. The shortfall is of a piece with weak \ac{tom} reasoning, and argues for training objectives that reward pragmatic inference from trajectories rather than fidelity to the trajectories themselves.

\paragraph{Qualitative analysis}
The failures are legible in individual episodes (\cref{fig:test_points}). Given demonstrations of ``wash fruits before cutting,'' end-to-end Gemini-3.1-Pro frequently omits the brief washing behavior and proceeds directly to cutting, responding to the salient objects in view rather than to the regularity they instantiate. The two-stage pipeline first commits to a constraint, \eg, ``the user goes to the sink before using the knife,'' and planning conditioned on that statement is far likelier to emit the requisite \texttt{toggle\_sink} action. Verbalization steadies the preference representation and keeps subtle constraints from being forgotten over a long horizon.

\section{Conclusion}

We have recast personalized embodied planning as a problem of pragmatic reasoning. \ac{pbp} evaluates implicit semantic alignment rather than instruction following, and reveals a substantial gap between what agents can do once a preference is given and what they can recover on their own. \ac{intu} shows the gap to be narrowable: an agent that states the preference before acting aligns with human intent more reliably than one imitating demonstrations directly. Language, on this evidence, is a workable bridge from observed behavior to personalized action, and pragmatics, language, and embodied intelligence meet on ground worth further study.

\paragraph{Limitations}
The benchmark is synthetic, and no simulator reproduces the full complexity of real environments. Preference labels are human-defined and may pass over finer variation in how people actually behave. We also supply the relevant demonstrations as context, whereas a deployed agent would first have to retrieve them from long-term memory, a step this work does not address.

\section{Conclusion}

In this work, we reframed the problem of personalized embodied planning as a pragmatic reasoning task. We introduced \ac{pbp}, a benchmark for evaluating implicit semantic alignment, and demonstrated that current agents exhibit a substantial gap between preference acquisition and preference-conditioned planning. Through experiments, we show that explicit verbalization of preferences enables agents to align with human intent more robustly than implicit visual imitation. We hope this work inspires further research into the intersection of pragmatics, language, and embodied intelligence.

\paragraph{Limitations} 
Our benchmark relies on synthetic data, and simulator environments cannot fully capture real-world complexity. Human-defined preference labels may also overlook subtle variations in user behavior. In addition, we assume that relevant demonstrations are provided as context; real-world agents must additionally retrieve relevant experiences from long-term memory.

\section*{Acknowledgments}

This work is supported in part by the National Natural Science Foundation of China (32595491,62376009), the Beijing Nova program, the NVIDIA Academic Grant Program using Spark and Thor, the State Key Lab of General AI at Peking University, the PKU-BingJi Joint Laboratory for Artificial Intelligence, the Wuhan Major Scientific and Technological Special Program (2025060902020304), the Hubei Embodied Intelligence Foundation Model Research and Development Program, and the National Comprehensive Experimental Base for Governance of Intelligent Society, Wuhan East Lake High-Tech Development Zone.

\bibliography{reference_header,references}
\clearpage

\appendix
\renewcommand\thefigure{A\arabic{figure}}
\setcounter{figure}{0}
\renewcommand\thetable{A\arabic{table}}
\setcounter{table}{0}
\renewcommand\theequation{A\arabic{equation}}
\setcounter{equation}{0}
\pagenumbering{arabic}%
\renewcommand*{\thepage}{A\arabic{page}}
\setcounter{footnote}{0}

\section{Dataset Card}

We follow the datasheet proposed in \citet{gebru2021datasheets} for documenting our proposed \ac{pbp}:
\begin{enumerate}
    \item \textbf{Motivation}
    \begin{enumerate}
    \item \textbf{For what purpose was the dataset created?}\\
    The benchmark was created to evaluate existing learning agents on their ability to understand and adapt to various human preferences. Specifically, it aims to test the agents' proficiency in few-shot learning from demonstrations, where they must respond to ambiguous task instructions and formulate adaptive task plans based on limited examples of user preferences. The benchmark is designed to highlight the challenges and gaps in current AI systems' capabilities in planning activities and abstracting human preferences, ultimately driving advancements towards developing more intelligent and personalized embodied agents.
    \item \textbf{Who created the dataset and on behalf of which entity?}\\
    N/A.
    \item \textbf{Who funded the creation of the dataset?}\\
    N/A.
    \item \textbf{Any other Comments?}\\
    None.
    \end{enumerate}
    \item \textbf{Composition}
    \begin{enumerate}
        \item \textbf{What do the instances that comprise the dataset represent?}\\
        Each instance contains an egocentric video of an agent's activity, its bird's-eye-view map of the position of the agent, and a frame-level textual annotation of the current action, as shown in \cref{fig:datapoint}. Additionally, we provide a rendered third-person view of the entire process.
        \item \textbf{How many instances are there in total?}\\
        15,000. 
        \item \textbf{Does the dataset contain all possible instances or is it a sample (not necessarily random) of instances from a larger set?}\\
        No. The dataset contains a set of demonstrations rendered within the simulator, The users can render more diverse instances if they want. We have provided the rendering instructions.
        \item \textbf{What data does each instance consist of?}\\
        The instances that comprise the benchmark represent various types of human preferences applied to different tasks within a realistic embodied scene. Each instance is designed to challenge the learning agents to understand and adapt to these preferences based on a few demonstration examples, reflecting the diverse and hierarchical nature of user preferences in real-world scenarios. See above for data details.
        \item \textbf{Is there a label or target associated with each instance?}\\
        Yes.
        \item \textbf{Is any information missing from individual instances?}\\
        No.
        \item \textbf{Are relationships between individual instances made explicit?}\\
        Yes.
        \item \textbf{Are there recommended data splits?}\\
        No.
        \item \textbf{Are there any errors, sources of noise, or redundancies in the dataset?}\\
        No.
        \item \textbf{Is the dataset self-contained, or does it link to or otherwise rely on external resources (\eg, websites, tweets, other datasets)?}\\
        Self-contained.
        \item \textbf{Does the dataset contain data that might be considered confidential (\eg, data that is protected by legal privilege or by doctor-patient confidentiality, data that includes the content of individuals' non-public communications)?}\\
        No.
        \item \textbf{Does the dataset contain data that, if viewed directly, might be offensive, insulting, threatening, or might otherwise cause anxiety?}\\
        No.
        \item \textbf{Does the dataset relate to people?}\\
        No.
        \item \textbf{Does the dataset identify any subpopulations (\eg, by age, gender)?}\\
        No.
        \item \textbf{Is it possible to identify individuals (\ie, one or more natural persons), either directly or indirectly (\ie, in combination with other data) from the dataset?}\\
        No.
        \item \textbf{Does the dataset contain data that might be considered sensitive in any way (\eg, data that reveals racial or ethnic origins, sexual orientations, religious beliefs, political opinions or union memberships, or locations; financial or health data; biometric or genetic data; forms of government identification, such as social security numbers; criminal history)?}\\
        No.
        \item \textbf{Any other comments?}\\
        None.
    \end{enumerate}
    \item \textbf{Collection Process}
    \begin{enumerate}
        \item \textbf{How was the data associated with each instance acquired?}\\
        We render \ac{pbp} using NVIDIA's Omniverse and OmniGibson simulation environment \citep{li2023behavior}. 
        \item \textbf{What mechanisms or procedures were used to collect the data (\eg, hardware apparatus or sensor, manual human curation, software program, software API)?}\\
        The data for each instance in the benchmark was acquired by sampling preferences from a predefined set and constructing tasks paired with a few demonstrations that shared high-level preferences but differed in specific objects and scenes. Each sampled preference was randomly assigned to one of the 50 scenes provided by OmniGibson, with relevant objects sampled within the scene. Egocentric observation and action sequences of an embodied agent were generated as the agent performed tasks guided by a rule-based planner using planning primitives like inverse kinematics for grasping and the A* algorithm for movement.  
        \item \textbf{If the dataset is a sample from a larger set, what was the sampling strategy (\eg, deterministic, probabilistic with specific sampling probabilities)?}\\
        N/A.
        \item \textbf{Who was involved in the data collection process (\eg, students, crowdworkers, contractors) and how were they compensated (\eg, how much were crowdworkers paid)?}\\
        N/A. 
        \item \textbf{Over what timeframe was the data collected?}\\
        N/A.
        \item \textbf{Were any ethical review processes conducted (\eg, by an institutional review board)?}\\
        The dataset raises no ethical concerns.
        \item \textbf{Does the dataset relate to people?}\\
        No.
        \item \textbf{Did you collect the data from the individuals in question directly, or obtain it via third parties or other sources (\eg, websites)?}\\
        N/A.
        \item \textbf{Were the individuals in question notified about the data collection?}\\
        N/A.
        \item \textbf{Did the individuals in question consent to the collection and use of their data?}\\
        N/A.
        \item \textbf{If consent was obtained, were the consenting individuals provided with a mechanism to revoke their consent in the future or for certain uses?}\\
        N/A.
        \item \textbf{Has an analysis of the potential impact of the dataset and its use on data subjects (\eg, a data protection impact analysis) been conducted?}\\
        Yes.
        \item \textbf{Any other comments?}\\
        None.
    \end{enumerate}
    \item \textbf{Preprocessing, Cleaning and Labeling}
    \begin{enumerate}
        \item \textbf{Was any preprocessing/cleaning/labeling of the data done (\eg, discretization or bucketing, tokenization, part-of-speech tagging, SIFT feature extraction, removal of instances, processing of missing values)?}\\
        N/A.
        \item \textbf{Was the ``raw'' data saved in addition to the preprocessed/cleaned/labeled data (\eg, to support unanticipated future uses)?}\\
        N/A.
        \item \textbf{Is the software used to preprocess/clean/label the instances available?}\\
        N/A.
        \item \textbf{Any other comments?}\\
        None.
    \end{enumerate}
    \item \textbf{Uses}
    \begin{enumerate}
        \item \textbf{Has the dataset been used for any tasks already?}\\
        No, the dataset is newly proposed by us.
        \item \textbf{Is there a repository that links to any or all papers or systems that use the dataset?}\\
        No, the dataset is new.
        \item \textbf{What (other) tasks could the dataset be used for?}\\
        This dataset could be used for research topics like embodied AI and human-computer interaction.
        \item \textbf{Is there anything about the composition of the dataset or the way it was collected and preprocessed/cleaned/labeled that might impact future uses?}\\
        N/A.
        \item \textbf{Are there tasks for which the dataset should not be used?}\\
        N/A.
        \item \textbf{Any other comments?}\\
        None.
    \end{enumerate}
    \item \textbf{Distribution}
    \begin{enumerate}
        \item \textbf{Will the dataset be distributed to third parties outside of the entity (\eg, company, institution, organization) on behalf of which the dataset was created?}\\
        No before it is made public.
        \item \textbf{How will the dataset be distributed (\eg, tarball on website, API, GitHub)?}\\
        On our project website upon acceptance.
        \item \textbf{When will the dataset be distributed?}\\
        Upon acceptance.
        \item \textbf{Will the dataset be distributed under a copyright or other intellectual property (IP) license, and/or under applicable terms of use (ToU)?}\\
        Under CC BY-NC \footnote{\url{https://creativecommons.org/licenses/by-nc/4.0/}} license.
        \item \textbf{Have any third parties imposed IP-based or other restrictions on the data associated with the instances?}\\
        No.
        \item \textbf{Do any export controls or other regulatory restrictions apply to the dataset or to individual instances?}\\
        No.
        \item \textbf{Any other comments?}\\
        None.
    \end{enumerate}
    \item \textbf{Maintenance}
    \begin{enumerate}
        \item \textbf{Who is supporting/hosting/maintaining the dataset?}\\
        The authors.
        \item \textbf{How can the owner/curator/manager of the dataset be contacted (\eg, email address)?}\\
        N/A.
        \item \textbf{Is there an erratum?}\\
        Future erratum will be released through the website.
        \item \textbf{Will the dataset be updated (\eg, to correct labeling errors, add new instances, delete instances')?}\\
        Yes.
        \item \textbf{If the dataset relates to people, are there applicable limits on the retention of the data associated with the instances (\eg, were individuals in question told that their data would be retained for a fixed period of time and then deleted)?}\\
        N/A. The dataset does not relate to people.
        \item \textbf{Will older versions of the dataset continue to be supported/hosted/maintained?}\\
        Yes.
        \item \textbf{If others want to extend/augment/build on/contribute to the dataset, is there a mechanism for them to do so?}\\
        Yes. We will release the source code as well as a licence on our project website after acceptance. 
        \item \textbf{Any other comments?}\\
        None.
    \end{enumerate}
\end{enumerate}

\section{Dataset Statistics}\label{sec:supp:dataset_statis}

Simulation lengths in the dataset range from 1 to 5 minutes, depending on the tasks recorded. And the videos are recorded at 30 fps.

\subsection{Preferences}

\begin{figure*}[ht!]
    \centering
    \small
    \begin{subfigure}{0.333\linewidth}
        \centering
        \includegraphics[trim={1.2cm 1.2cm 1.2cm 1.2cm}, clip, width=\linewidth]{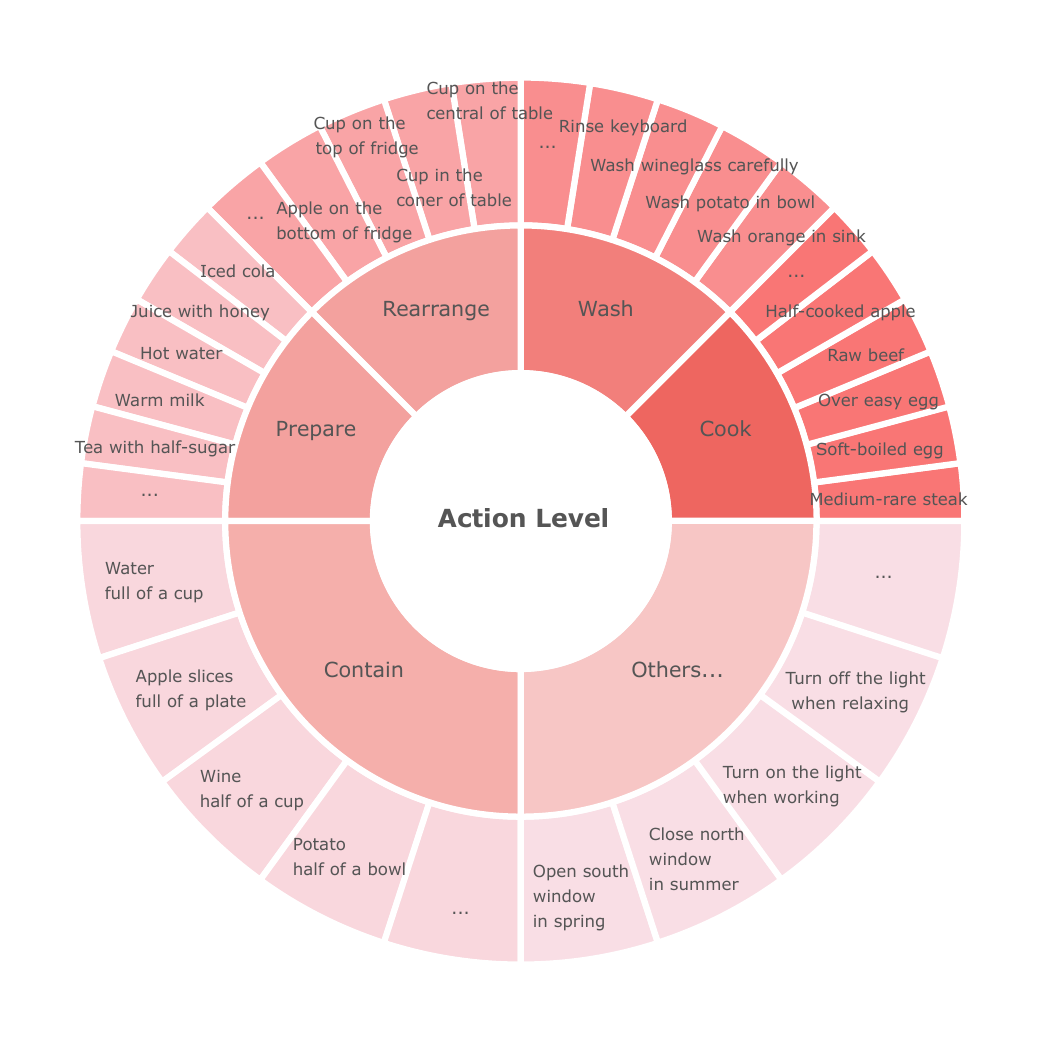}
        \caption{Action level}
    \end{subfigure}%
    \hfill
    \begin{subfigure}{0.333\linewidth}
        \centering
        \includegraphics[trim={1.2cm 1.2cm 1.2cm 1.2cm}, clip, width=\linewidth]{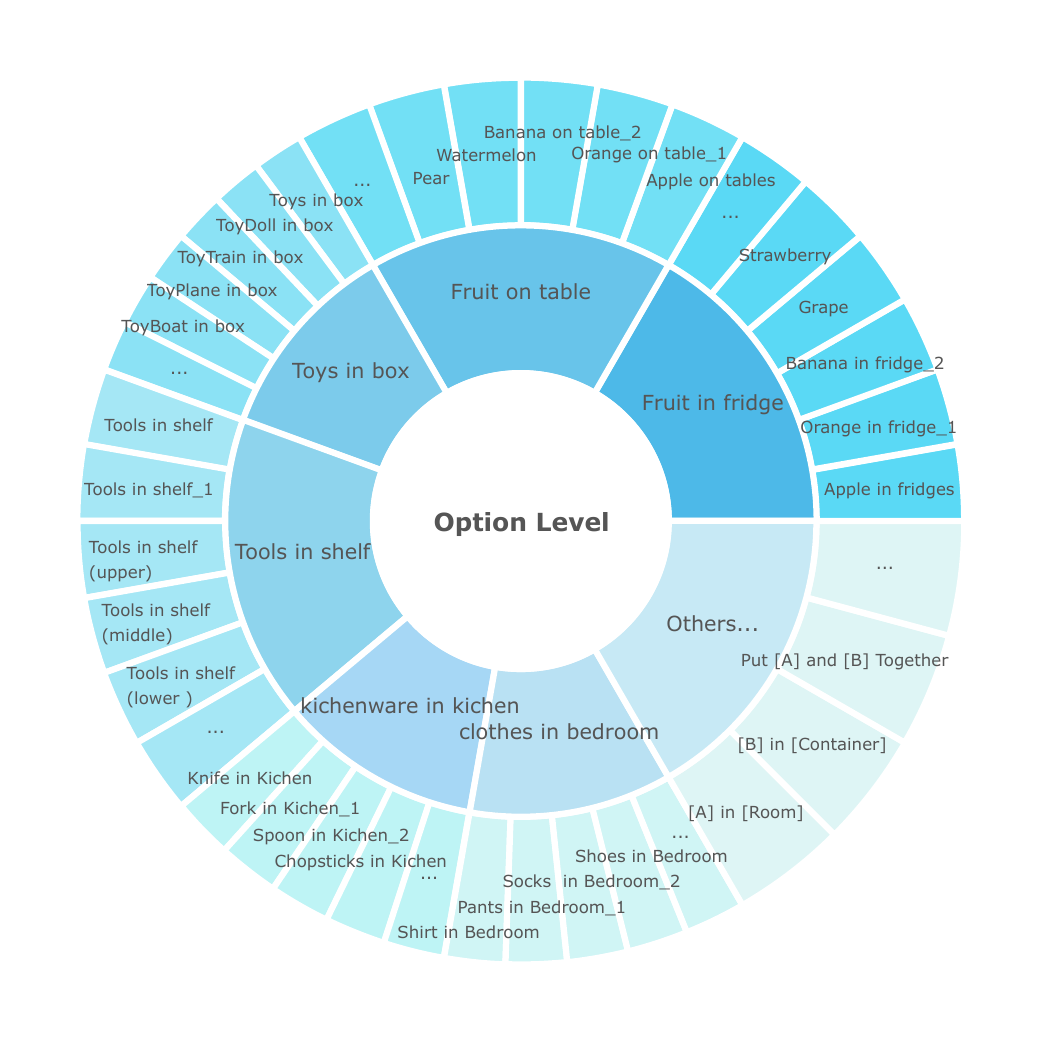}
        \caption{Option level}
    \end{subfigure}%
    \hfill
    \begin{subfigure}{0.333\linewidth}
        \centering
        \includegraphics[trim={1.2cm 1.2cm 1.2cm 1.2cm}, clip, width=\linewidth]{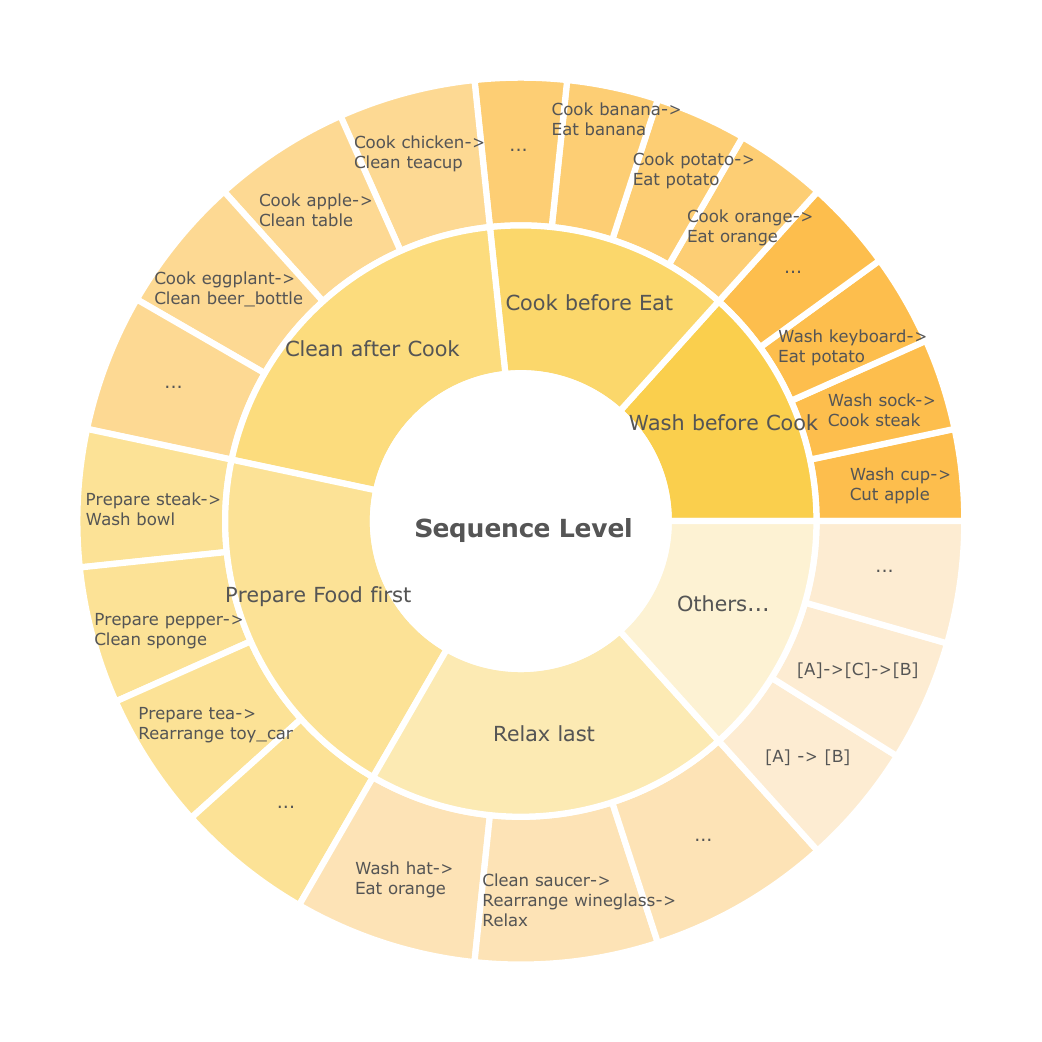}
        \caption{Sequence level}
    \end{subfigure}%
    \caption{\textbf{Hierarchical organization of user preferences.} Our framework organizes preferences in a three-tiered structure, visualized through sunburst diagrams: (a) Action level captures fine-grained execution details within specific tasks, from quantity preferences in ``Contain'' (\eg, ``half a cup'' \vs ``full cup'') to environmental controls (\eg, lighting and window operations). (b) Option level represents spatial preferences for object categories, encoding both storage decisions (\eg, table \vs fridge for fruits) and organizational choices (\eg, shelf levels and boxes for tools/toys). (c) Sequence level defines temporal relationships between tasks, encompassing both basic preparation sequences (\eg, ``Prepare Food first'') and conditional orderings (\eg, ``Clean after Cook,'' ``[A]->[B]''). Each diagram's hierarchical structure branches from general categories to specific instances, revealing detailed preference patterns upon closer inspection. (Vector graphics; zoom in for details.)}
    \label{fig:statistics}
\end{figure*}

See \cref{tab:dataset_statistics} for the preference statistics in \ac{pbp}. See \cref{fig:statistics} for the hierarchical organization of user preferences. 
\begin{table}[ht!]
    \setlength{\tabcolsep}{3pt}
    \caption{\textbf{Dataset Statistics in \ac{pbp}.}}
    \centering
    \small
    \rowcolors{2}{LightCyan1}{}
    \begin{tabular}{cccc} \toprule
          & \textbf{Action} & \textbf{Option} & \textbf{Sequence}\\ \midrule
         \textbf{Preference Num} & 75 & 135 & 80\\ 
         \textbf{Episode Num} & 5000 & 5000 & 5000\\
         \textbf{Sub-task Num} & 1 & 2-3 & 2-3 \\ \bottomrule
    \end{tabular}
    \label{tab:dataset_statistics}
\end{table}

\subsection{Actions}\label{sec:supp:statistics}

See \cref{tab:action_statistics} for the action statistics in \ac{pbp}. We implement 17 action primitives in \ac{pbp} to assist with model planning and dataset rendering. These action primitives have parameters that simplify tasks and are considered the lowest-level actions. Each sub-task contains 8 to 20 such lowest-level actions. Generally, most of these actions consist of two parts: the robot movement part and the arm (gripper) execution part. For robot movement, we use the A* algorithm to find paths and avoid collisions. We build a connection map during scene initialization for navigation, taking the robot's width into consideration. For the arm (gripper) execution, we primarily use the \ac{ik} algorithm to compute arm movements. However, since \ac{ik} cannot handle complex tasks, such as picking objects from the fridge, we also leverage the \ac{ompl} planner \citep{sucan2012open} with forward planning to assist in planning the arm positions.

\subsection{More Dataset Details and Discussion}
\paragraph{Dataset production} In summary, we follow the order of ``sample preference - sample scene - sample objects to be manipulated - generate actions guided by a rule-based planner''.
\paragraph{Length and FPS of the simulations} The length of the simulations ranges from 1 to 5 minutes, depending on the tasks recorded. The videos are recorded at 30 fps. 
\paragraph{Actions contained in each simulation} The number of actions in simulations varies among different preference levels. There is 1 subtask for action-level, 2-3 subtasks for option-level, and 2-3 subtasks for sequence-level preferences. Each subtask contains 8-20 actions. 
\paragraph{Scenes and rooms} Each scene contains various types of rooms. The main differences between scenes are the type, number, and layout of both rooms and furniture. Additionally, each room may contain different objects and have unique layouts. Details of the scenes and rooms can be found in Omnigibson's official documentation (https://behavior.stanford.edu/omnigibson/), as we directly adopt these scenes from the open-sourced project. 
\paragraph{290 preference types} Considering that preferences in household activities are not only multi-dimensional but also hierarchical, we first define a hierarchy of preferences from the perspective of how things happen in a life scenario, that is, from each specific action to a sub-task consisting of several actions, and then to the sequence combining these sub-tasks. The next step is to expand each level with typical tasks and actions.
\paragraph{The egocentric view} Collecting both egocentric observations and third-person views is feasible in PbP or similar environments built on simulators like iGibson. However, in real-world scenarios, it is generally easier to gather egocentric observations of human daily activities, as these can be efficiently captured through wearable devices. Additionally, there are numerous egocentric-view datasets available, such as Ego4D\citep{grauman2022ego4d}, which further facilitate this approach. While third-person views can provide a different perspective, they often encounter issues such as occlusion. Although research based on third-person views is essential for applications involving real robots, focusing on egocentric views in the current work allows for a more straightforward exploration of preference learning and planning. Nevertheless, third-person view data can be obtained by integrating additional cameras, as outlined in our provided code.
\paragraph{Action ground truth} In experiments involving vision input, we do not explicitly provide the action sequence of the user. In the symbolic-based experiment, we provide the action sequence to reduce the perception cost to concentrate more effectively on the inference and planning aspects of the study. 

\begin{table*}[ht!]
    \small
    \centering
    \setlength{\tabcolsep}{3pt}
    \caption{Action Primitives in \ac{pbp}.}
    \rowcolors{2}{LightCyan1}{}
    \begin{tabular}{cc} \toprule
        \textbf{Action List} & \textbf{Explanation}\\ \midrule
        \textbf{Move\_to\_[]} & Move to a specified location, or a specified room, or a specified object\\ 
        \textbf{Rotate\_to\_[]} & Rotate to a specified orientation or a specified object\\ 
        \textbf{Pick\_[]} & Pick up an object using the gripper, \eg, ``Pick\_apple''\\ 
        \textbf{Place\_[]} & Place an object at a location, \eg, ``Place\_apple\_on\_table''\\ 

        \textbf{Fill\_[]\_with\_[]} & Fill a container with a substance, \eg, ``Fill\_glass\_with\_water''\\ 
        \textbf{Pour\_[]} & Pour a substance from a container, \eg, ``Pour\_milk''\\ 
        \textbf{Open\_[]} & Open an object, \eg, ``Open\_door''\\ 
        \textbf{Close\_[]} & Close an object, \eg, ``Close\_fridge''\\ 
        \textbf{Cut\_[]} & Cut an object, \eg, ``Cut\_carrot''\\ 
        \textbf{Cook\_[]} & Cook an item, \eg, ``Cook\_pasta''\\ 
        \textbf{Wash\_[]} & Wash an object, \eg, ``Wash\_dishes''\\ 
        \textbf{Clean\_[]} & Clean a surface or object, \eg, ``Clean\_counter''\\ 
        \textbf{Cover\_[]} & Cover an object, \eg, ``Cover\_bowl''\\ 
        \textbf{Uncover\_[]} & Uncover an object, \eg, ``Uncover\_bowl''\\ 
        \textbf{Toggle\_on\_[]} & Turn on a device, \eg, ``Toggle\_on\_light''\\ 
        \textbf{Toggle\_off\_[]} & Turn off a device, \eg, ``Toggle\_off\_stove''\\ 
        \textbf{Wait\_[]} & Wait some time\\
        \bottomrule
    \end{tabular}
    \label{tab:action_statistics}
\end{table*}

\section{Experiment Details}\label{sec:supp:experiment}

\subsection{Archived Experimental Results}\label{sec:supp:archived}

The following tables record experimental results from previous model 
iterations that have been superseded in the current version of this work. 
These results are retained for completeness and archival reference.

\begin{table*}[ht!]
    \centering
    \small
    \caption{\textbf{Archived Results --- Levenshtein Distance (Superseded Models).} 
    Results for GPT-4V (Video), DeepSeek-R1, GPT-4.1, and GPT-5.2 from \cref{tab:results_levenshtein_distance}. 
    These have been superseded by updated experiments with Gemini 3 Pro, DeepSeek-v4-pro, Claude Opus 4.7, and GPT-5.5.}
    \label{tab:archived_levenshtein}
    \setlength{\tabcolsep}{3pt}
    \begin{tabular}{cccccc}
        \toprule
        & \multicolumn{5}{c}{\textbf{DEPRECATED MODELS}} \\
        \cmidrule(r){2-6}
        & GPT-4V & Llama3-8B & DeepSeek-R1 & GPT-4.1 & GPT-5.2 \\
        \midrule
        \multicolumn{6}{c}{\textbf{Option Level}} \\
        End-to-end 
            & 15.63$\pm$2.31 & 14.74$\pm$3.21 & 8.73$\pm$3.03  & 7.42$\pm$2.67  & 7.41$\pm$2.78 \\
        Two-stage 
            & 8.37$\pm$2.19  & 9.67$\pm$5.16  & 3.19$\pm$2.19  & 2.26$\pm$2.03  & 2.18$\pm$2.33 \\
        Second-stage (gt) 
            & 1.26$\pm$2.55  & 8.22$\pm$5.58  & 1.76$\pm$1.89  & 0.15$\pm$2.85  & 0.17$\pm$2.46 \\
        \midrule
        \multicolumn{6}{c}{\textbf{Sequence Level}} \\
        End-to-end 
            & 33.75$\pm$11.15 & 31.79$\pm$7.32 & 28.72$\pm$4.12 & 26.48$\pm$3.25 & 26.00$\pm$3.34 \\
        Two-stage 
            & 27.52$\pm$9.48  & 25.46$\pm$5.93 & 18.61$\pm$3.25 & 14.19$\pm$3.01 & 12.61$\pm$2.77 \\
        Second-stage (gt) 
            & 11.36$\pm$8.05  & 19.02$\pm$7.10 & 14.10$\pm$3.76 & 10.31$\pm$2.98 & 10.17$\pm$3.32 \\
        \midrule
        \multicolumn{6}{c}{\textbf{Overall}} \\
        End-to-end 
            & 24.69 & 23.26 & 18.72 & 16.95 & 16.70 \\
        Two-stage 
            & 17.94 & 17.56 & 10.90 & 8.22  & 7.40  \\
        Second-stage (gt) 
            & 6.31  & 13.62 & 7.93  & 5.23  & 5.17  \\
        \bottomrule
    \end{tabular}%
\end{table*}

\begin{table*}[ht!]
    \centering
    \small
    \caption{\textbf{Archived Results --- Preference Inference Accuracy (Superseded Models).} 
    Results for GPT-4V (Video), DeepSeek-R1, GPT-4.1, and GPT-5.2 from \cref{tab:results_combined}. 
    These have been superseded by updated experiments with Gemini 3 Pro, DeepSeek-v4-pro, Claude Opus 4.7, and GPT-5.5.}
    \label{tab:archived_accuracy}
    \setlength{\tabcolsep}{3pt}
    \begin{tabular}{cccccc}
        \toprule
        & \multicolumn{5}{c}{\textbf{DEPRECATED MODELS}} \\
        \cmidrule(r){2-6}
        & GPT-4V & Llama3-8B & DeepSeek-R1 & GPT-4.1 & GPT-5.2 \\
        \midrule
        \multicolumn{6}{c}{\textbf{Few-shot (With Context Demonstrations)}} \\
        Option Level 
            & 48.48 & 72.98 & 86.02 & 88.91 & 89.78 \\
        Sequence Level 
            & 37.50 & 67.18 & 71.21 & 70.28 & 73.34 \\
        Overall 
            & 42.99 & 70.08 & 78.62 & 79.60 & 81.56 \\
        \midrule
        \multicolumn{6}{c}{\textbf{Ablative (Zero-shot / No Context)}} \\
        Option Level 
            & 29.42 & 39.50 & 78.19 & 75.29 & 78.24 \\
        Sequence Level 
            & 0.00  & 6.25  & 15.29 & 14.11 & 18.44 \\
        Overall 
            & 14.71 & 22.88 & 46.74 & 44.70 & 48.34 \\
        \bottomrule
    \end{tabular}%
\end{table*}

\begin{table*}[ht!]
    \centering
    \small
    \caption{\textbf{Archived Results --- Generalization (Superseded Models).} 
    Results for GPT-4V, DeepSeek-R1, GPT-4.1, and GPT-5.2 from \cref{table:results_generalization}. 
    These have been superseded by updated experiments with Gemini 3 Pro, DeepSeek-v4-pro, Claude Opus 4.7, and GPT-5.5.}
    \label{tab:archived_generalization}
    \begin{tabular}{cccccc}
        \toprule
        & GPT-4V & DeepSeek-R1 & GPT-4.1 & GPT-5.2 \\
        \midrule
        Option             & 53.24 & 86.02 & 88.91 & 89.78 \\
        Option \textit{gen} & 48.48 & 84.98 & 86.57 & 86.57 \\
        Sequence            & 39.42 & 71.21 & 70.28 & 73.34 \\
        Sequence \textit{gen} & 37.50 & 70.16 & 69.28 & 69.28 \\
        \bottomrule
    \end{tabular}%
\end{table*}

\subsection{Case Study}

We also provide a case with preference \textit{Put fruit on the bed} in the following table \cref{supp_table_cases}. We present a simplified version of the demonstrations, where all video outputs have been translated into symbol-based action sequences for ease of understanding. Video-based models such as LLaVA-Next and GPT-4V struggle with comprehending preferences and tend to replicate certain action patterns from the video demonstration, such as ``move to'' and ``pick up''. Llama3 demonstrates a partial understanding and execution of the preference. It correctly moves to each fruit (grape, banana), picks them up, and places them on the bed. However, it also interacts with the pencil and places it on the bed, which is not required by the preference. Ideally, the pencil should be placed on the table, similar to the pen. On the other hand, GPT-4.1(Symbol) accurately interacts with the grape and banana by moving to each fruit, picking it up, and placing it on the bed. This demonstrates a better understanding and execution of the preference compared to the other models.

\begin{table*}[ht!]
    \small
    \centering
    \setlength{\tabcolsep}{3pt}
    \begin{tabular}{cp{3cm}p{2cm}p{8cm}}
        \toprule
        \textbf{Case} & \textbf{ID} & \textbf{Scene} & \textbf{Actions}\\
        \midrule
        \textbf{Context} & Demo 1 & Beechwood & move to orange, pick up orange, move to bed with orange, place orange on bed, move to apple, pick up apple, move to bed with apple, place apple on bed, move to dvd, pick up dvd, move to table with dvd, place dvd on table. \\
         & Demo 2 & Rs & move to banana, pick up banana, move to bed with banana, place banana on bed, move to club sandwich, pick up club sandwich, move to countertop with club sandwich, place club sandwich on countertop, move to orange, pick up orange, move to countertop with orange, move to bed with orange, place orange on bed. \\
        & Demo 3 & Merom & move to pen, pick up pen, move to table with pen, place pen, move to strawberry, pick up strawberry, move with to bed with strawberry, place strawberry on bed, move to textbook, pick up textbook, move with textbook, move to bookshelf with textbook, place textbook in bookshelf. \\\midrule
        \textbf{Test Scene} & & Beechwood & grape, pencil, banana \\\midrule
        \textbf{Answer} & LLaVA-Next & Beechwood & move to grape, move to bed with grape,       move to pencil, pick up pencil, move to banana, pick up banana \\
         & GPT-4V & Beechwood & move to grape, pick up grape, move to bed with grape, place grape on bed \\
         & Llama3 & Beechwood & move to grape, pick up grape, move to bed with grape, place grape on bed, move to pencil, pick up pencil, move to bed with pencil, place pencil on bed, move to banana, pick up banana, move to bed with banana, place banana on bed \\
         & GPT-4.1(Symbol) & Beechwood & move to grape, pick up grape, move to bed with grape, place grape on bed, move to banana, pick up banana, move to bed with banana, place banana on bed \\
        \bottomrule 
    \end{tabular}
    \label{supp_table_cases}
    \caption{Case Study with preference \textit{Put fruit on the bed}.}
\end{table*}

\section{Baseline Details}\label{sec:supp:baseline}

We evaluate two families of models corresponding to two practically relevant settings.
\textbf{Video-based models} operate directly on egocentric observations and therefore reflect the full challenge of extracting user preferences from raw behavioral evidence.
\textbf{Symbol-based models} receive textual action traces and scene descriptions, removing most perceptual uncertainty and providing a controlled setting for diagnosing preference reasoning and downstream planning.
Across both settings, we use a common two-stage evaluation protocol whenever supported by the model: preference prediction followed by preference-conditioned planning.

\subsection{Video-Based Models}

\paragraph{LLaVA-NeXT}
We evaluate LLaVA-NeXT-Video-7B-DPO \citep{liu2024improved}, an open-source multimodal model with native video support.
Following the official implementation, we use the \texttt{vicuna\_v1} conversation template, sample 32 frames per video, and use a spatial pooling stride of 2.
LLaVA-NeXT represents a practical open-source setting in which the agent must infer preferences directly from temporally extended egocentric observations.

\paragraph{EILEV}
We additionally evaluate EILEV \citep{yu2023efficient}, a multimodal model specifically designed for few-shot in-context learning from embodied videos.
Our implementation uses OPT-2.7B \citep{zhang2022opt} as the language backbone.
Because EILEV is pretrained on Ego4D \citep{grauman2022ego4d}, it provides a useful baseline for testing whether egocentric pretraining improves preference inference from behavioral demonstrations.
Unlike the other models, EILEV expects interleaved video--text examples in a specific in-context format; we therefore preserve its native demonstration ordering while keeping the semantic content of the prompt unchanged.

\paragraph{GPT-4V and Gemini-3.1-Pro}
We use GPT-4V \citep{achiam2023gpt} as an archived closed-source multimodal baseline and Gemini-3.1-Pro as the stronger model in the main experiments.
For GPT-4V, we use the Azure OpenAI API with \texttt{gpt-4-turbo-2024-04-09} and a temperature of 0.05.
Because of image-count limitations, each video is temporally subsampled to a compact set of frames.
For Gemini-3.1-Pro, we use the Google GenAI API with model id \texttt{gemini-3.1-pro-preview}, also with temperature 0.05.
We extract keyframes at 1 Hz and interleave them with text delimiters that preserve demonstration boundaries.
These models approximate a practical deployment scenario in which a general-purpose multimodal foundation model observes user behavior directly and must recover preference-relevant semantics without access to symbolic action annotations.

\subsection{Symbol-Based Models}

To isolate preference reasoning from visual perception, we evaluate LLMs on symbolic versions of the same demonstrations.
Each video is replaced by a textual scene description and action trace.
This setting is not intended to represent a more realistic interface; instead, it serves as a diagnostic upper-bound setting in which the relevant behavioral events are explicitly observable.

\paragraph{Llama3-8B}
We evaluate Meta-Llama-3-8B-Instruct \citep{touvron2023llama} using the official implementation with temperature 0.05.
Llama3-8B provides a relatively lightweight open-source language baseline for preference inference and planning from symbolic trajectories.

\paragraph{DeepSeek-R1 and GPT-4.1}
DeepSeek-R1 and GPT-4.1 are retained as archived symbolic baselines.
For DeepSeek-R1, we use a self-hosted 671B model with temperature 0.05 and suppress the explicit reasoning trace using an empty \texttt{<think>} block, as this setting did not improve performance in our experiments.
For GPT-4.1, we use \texttt{gpt-4.1-2025-04-14} with temperature 0.05.
Their archived results are reported in \cref{sec:supp:archived}.

\paragraph{GPT-5.5, Claude-Opus-4.7, and DeepSeek-v4-pro}
The main symbolic experiments use GPT-5.5, Claude-Opus-4.7, and DeepSeek-v4-pro.
GPT-5.5 is accessed through the OpenAI API, Claude-Opus-4.7 through the Anthropic API, and DeepSeek-v4-pro through a self-hosted deployment.
All are evaluated with temperature 0.05 using the same symbolic-input protocol.
Because these models receive ground-truth action traces rather than pixels, their performance primarily reflects preference inference and planning rather than visual recognition.

\subsection{Prompting Protocol}

We use two preference-inference prompt formats for different evaluation purposes.

\paragraph{Setting A: Query-Conditioned Preference Inference}
This is the main setting used throughout our experiments.
The model receives a set of unlabeled behavioral demonstrations together with the current task context.
The demonstrations may reflect different aspects of the user's behavior and need not correspond to a single identical preference.
The model must identify which behavioral regularities are relevant to the current query and infer the preference that should guide the present task.

\begin{Verbatim}[fontsize=\footnotesize, breaklines=true]
"Preference Inference"

You are a robot assistant that infers a user's preference from prior behavior.

All possible preferences are:
{ALL POSSIBLE PREFERENCES}

Here are several previous behavioral demonstrations:
[DEMO_1]
[DEMO_2]
[DEMO_3]

The current instruction is:
[INSTRUCTION]

The current scene is:
[SCENE_DESCRIPTION]

Based on the user's previous behavior, infer the preference
that is relevant to the current task.

Question:
What preference should be applied in the current situation?
Choose from the preferences listed above.
\end{Verbatim}

We use this prompt for the main personalized preference-acquisition setting, where the model must infer the latent preference relevant to the current task by jointly reasoning over unlabeled behavioral history, the instruction, and the current environment.

\paragraph{Setting B: Labeled Few-Shot Preference Recognition}
For completeness, we provide an alternative labeled few-shot prompt template
for preference recognition under explicit in-context supervision.
This protocol is not used for the main experimental results reported in this paper.

\begin{Verbatim}[fontsize=\footnotesize, breaklines=true]
"Few-Shot Preference Recognition"

You are a robot assistant that recognizes user preferences from behavior.

All possible preferences are:
{ALL POSSIBLE PREFERENCES}

Here are several labeled behavioral examples:
[DEMO_1] The preference is [PREFERENCE_1]
[DEMO_2] The preference is [PREFERENCE_2]
[DEMO_3] The preference is [PREFERENCE_3]

Now observe the following unlabeled query behavior:
[TEST_CASE]

Question:
What preference is expressed by the query behavior?
Choose from the preferences listed above.
\end{Verbatim}

\paragraph{Stage Two: Preference-Conditioned Planning}
To evaluate whether explicit preference representations improve planning, we provide the inferred preference directly to the planner:

\begin{Verbatim}[fontsize=\footnotesize, breaklines=true]
"Stage Two / Preference-Conditioned Planning"

You are a robot assistant that generates an action sequence
consistent with the user's preference.

The inferred user preference is:
[PREDICTED_PREFERENCE]

The current instruction is:
[INSTRUCTION]

The current scene is:
[SCENE_DESCRIPTION]

All possible actions are:
{ALL POSSIBLE ACTIONS}

Generate an action sequence that satisfies the instruction
while respecting the inferred preference:
\end{Verbatim}

For the oracle \textit{Second-stage (gt)} condition, we use the same template but replace
\texttt{[PREDICTED\_PREFERENCE]} with \texttt{[GROUND\_TRUTH\_PREFERENCE]}.
This controlled substitution allows us to measure how much planning error is attributable to preference acquisition rather than downstream action generation.

\paragraph{End-to-End Planning}
The end-to-end condition removes the explicit preference representation and asks the model to generate a plan directly from behavioral context:

\begin{Verbatim}[fontsize=\footnotesize, breaklines=true]
"End-to-End Planning"

You are a robot assistant that generates an action sequence
based on the observed user behavior.

Behavioral context:
[DEMONSTRATIONS]

The current instruction is:
[INSTRUCTION]

The current scene is:
[SCENE_DESCRIPTION]

All possible actions are:
{ALL POSSIBLE ACTIONS}

Generate the action sequence:
\end{Verbatim}

The main evaluation therefore consists of three complementary components: \textit{unlabeled preference induction},
\textit{preference-conditioned planning}, and \textit{end-to-end planning}. Together, they ask whether the model can infer a latent preference from behavior, whether it can plan effectively once that preference is explicit, and whether it can solve both problems jointly without an explicit intermediate representation.

\subsection{Implementation Summary}

For reproducibility, \cref{tab:model_versions} summarizes the models used in the main experiments.
Archived models are retained only for historical comparison.

\begin{table}[ht!]
    \small
    \centering
    \setlength{\tabcolsep}{4pt}
    \caption{Model versions and access methods used in the main experiments.}
    \rowcolors{2}{LightCyan1}{}
    \resizebox{\linewidth}{!}{%
    \begin{tabular}{lll}
        \toprule
        \textbf{Model} & \textbf{Version / ID} & \textbf{Access} \\
        \midrule
        ViViT           & official Scenic implementation & local \\
        LLaVA-NeXT      & LLaVA-NeXT-Video-7B-DPO        & local \\
        EILEV           & EILEV + OPT-2.7B                & local \\
        Gemini-3.1-Pro  & \texttt{gemini-3.1-pro-preview} & Google API \\
        Llama3-8B       & Meta-Llama-3-8B-Instruct        & local \\
        DeepSeek-v4-pro & deepseek-v4-pro                  & self-hosted \\
        Claude-Opus-4.7 & \texttt{claude-opus-4-7}        & Anthropic API \\
        GPT-5.5         & \texttt{gpt-5.5}                & OpenAI API \\
        \bottomrule
    \end{tabular}}
    \label{tab:model_versions}
\end{table}

\section{Explicit Preference Reasoning}

To distinguish whether the improvement of \ac{intu} comes from explicitly reasoning about latent preferences or from exposing the preference as a separate intermediate representation, we introduce a Preference-aware End-to-End baseline. Unlike the vanilla End-to-End setting, which directly generates actions from behavioral demonstrations without being prompted to consider latent preferences, this baseline explicitly asks the model to infer the user's preference internally and directly produce the final action sequence without outputting the intermediate preference. We compare this setting with Two-stage Verbalization, where the inferred preference is explicitly verbalized before planning. This experiment allows us to separate the benefit of preference reasoning from the additional advantages of an interpretable and controllable intermediate representation.

\begin{table}[ht!]
    \small
    \centering
    \caption{\textbf{Effect of Explicit Preference Reasoning.}
    We compare vanilla End-to-End planning, Preference-aware End-to-End reasoning (P-aware), and Two-stage Verbalization.
    Preference-aware End-to-End asks the model to infer preferences internally before generating actions, while Two-stage Verbalization explicitly exposes the inferred preference as an intermediate representation. Lower Levenshtein distance indicates better alignment.}
    \rowcolors{2}{LightCyan1}{}
    \resizebox{\linewidth}{!}{%
    \begin{tabular}{lccc}
        \toprule
        \textbf{Model} 
        & \textbf{Vanilla} $\downarrow$
        & \textbf{P-aware} $\downarrow$
        & \textbf{Two-stage} $\downarrow$ \\
        \midrule
        \multicolumn{4}{c}{\textbf{Option Level}} \\
        Llama3-8B 
        & 14.74 $\pm$ 3.21 
        & 9.24 $\pm$ 4.11 
        & 9.67 $\pm$ 5.16 \\
        Gemini-3.1-Pro 
        & 8.72 $\pm$ 2.26 
        & 4.56 $\pm$ 1.96 
        & 4.28 $\pm$ 1.76 \\
        \midrule
        \multicolumn{4}{c}{\textbf{Sequence Level}} \\
        Llama3-8B 
        & 31.79 $\pm$ 7.32 
        & 28.13 $\pm$ 5.34 
        & 25.46 $\pm$ 5.93 \\
        Gemini-3.1-Pro 
        & 24.51 $\pm$ 4.15 
        & 15.73 $\pm$ 3.55 
        & 12.10 $\pm$ 2.85 \\
        \midrule
        \multicolumn{4}{c}{\textbf{Overall}} \\
        Llama3-8B 
        & 23.26 
        & 18.68 
        & 17.56 \\
        Gemini-3.1-Pro 
        & 16.62 
        & 10.14 
        & 8.19 \\
        \bottomrule
    \end{tabular}}
    \label{tab:preference_aware_baseline}
\end{table}

\end{document}